\documentclass[]{fairmeta}

\usepackage{tabularx}
\usepackage{textcomp}
\usepackage{url}
\usepackage{verbatim}
\usepackage{titlesec}
\usepackage{adjustbox}
\usepackage{multirow}
\usepackage{pifont}
\usepackage{tikz}
\usepackage{comment}
\usepackage{amsmath,amssymb} %
\usepackage{colortbl}  %
\usepackage{booktabs} 
\usepackage{hyperref}
\usepackage{graphicx}    %
\usepackage{subcaption} 
\definecolor{headerpurple}{HTML}{d8d2fc}
\definecolor{tablepurple}{HTML}{E4DDF9}
\RequirePackage{xspace}
\makeatletter
\DeclareRobustCommand\onedot{\futurelet\@let@token\@onedot}
\def\@onedot{\ifx\@let@token.\else.\null\fi\xspace}
\usepackage[most]{tcolorbox}
\usepackage{xcolor}
\usepackage{enumitem}
\usepackage{makecell}
\usepackage{mdframed}
\definecolor{keywordcolor}{RGB}{178,34,34} %
\usepackage{siunitx}

\newcolumntype{C}{>{\centering\arraybackslash}X}

\newcommand{\huggingface}{\raisebox{-1.5pt}{\includegraphics[height=1.05em]{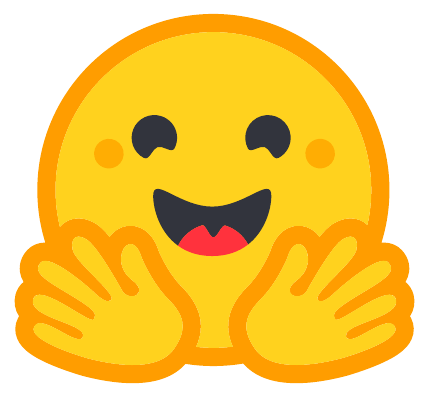}}\xspace}
\newcommand{\github}{\raisebox{-1.5pt}{\includegraphics[height=1.05em]{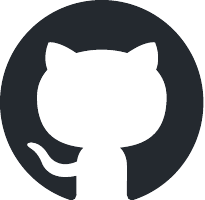}}\xspace}
\newcommand{\homepage}{\raisebox{-1.5pt}{\includegraphics[height=1.05em]{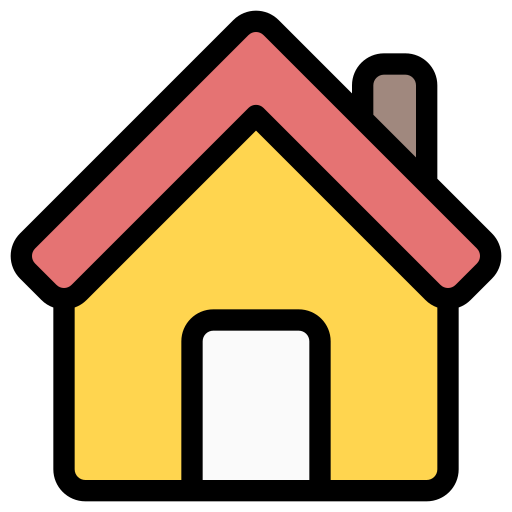}}\xspace}
\newcommand{\modelscope}{\raisebox{-1pt}{\includegraphics[height=0.95em]{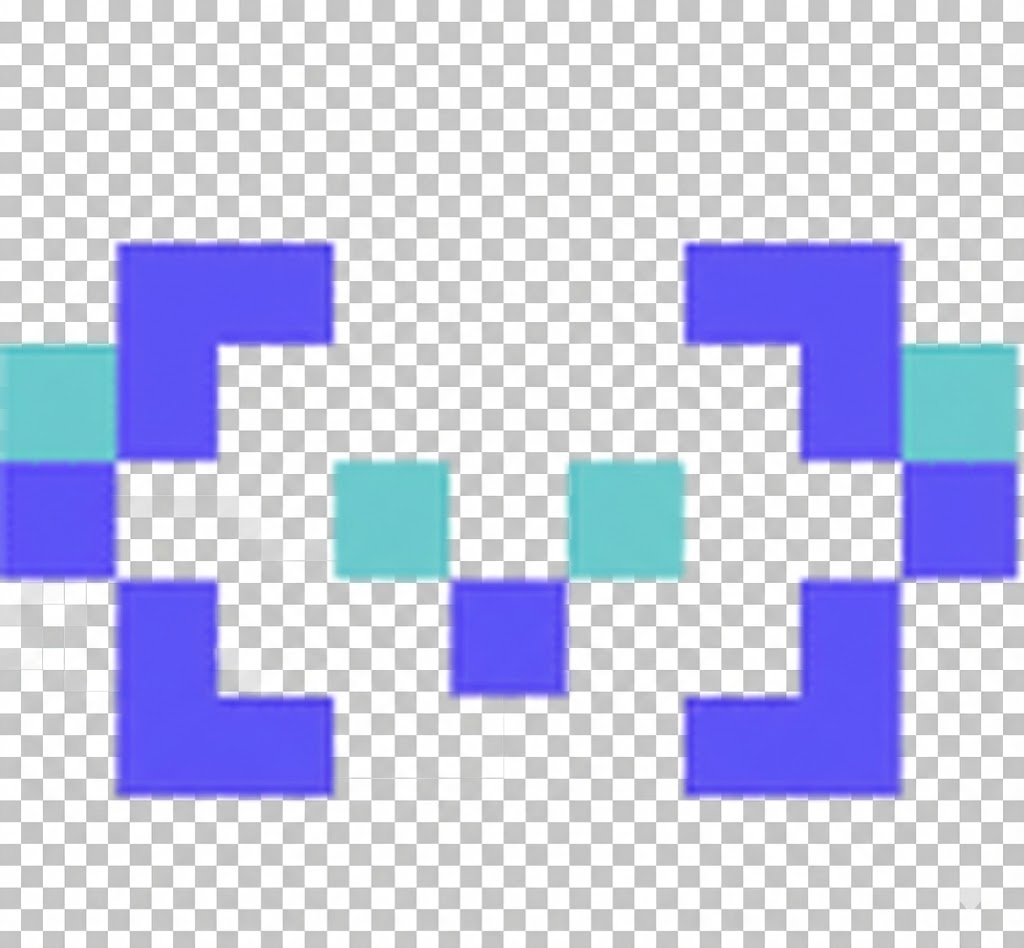}}\xspace}

\makeatother

\definecolor{adptorange}{RGB}{248, 205, 172}
\definecolor{cmpblue}{RGB}{189, 215, 238}

\definecolor{our_red}{RGB}{232,157,160}
\definecolor{our_blue}{RGB}{136,206,230}
\definecolor{our_orange}{RGB}{246,200,168}
\definecolor{our_green}{RGB}{178,211,164}

\definecolor{attn_code0}{RGB}{247,215,200}
\definecolor{attn_code1}{RGB}{238,169,139}
\definecolor{mlp_code0}{RGB}{204,201,221}
\definecolor{mlp_code1}{RGB}{102,95,153}

\definecolor{token_blue}{RGB}{84, 120, 140}

\usepackage{bbding}       %
\usepackage{fontawesome}
\usepackage{xspace}

\usepackage{float}
\newcommand{\our}{{RynnVLA-Latent}\xspace}
\newcommand{\lam}{{RynnLAM}\xspace}

\newlength\savewidth

\newcolumntype{x}[1]{>{\centering\arraybackslash}p{#1pt}}
\newcolumntype{y}[1]{>{\raggedright\arraybackslash}p{#1pt}}
\newcolumntype{z}[1]{>{\raggedleft\arraybackslash}p{#1pt}}

\renewcommand{\paragraph}[1]{\vspace{1mm}\noindent\textbf{#1}}

\usepackage{algorithm}
\usepackage{listings}

\definecolor{codeblue}{rgb}{0.25, 0.5, 0.5}
\definecolor{codekw}{rgb}{0.35, 0.35, 0.75}
\lstdefinestyle{Pytorch}{
    language = Python,
    backgroundcolor = \color{white},
    basicstyle = \fontsize{9pt}{8pt}\selectfont\ttfamily\bfseries,
    columns = fullflexible,
    aboveskip=1pt,
    belowskip=1pt,
    breaklines = true,
    captionpos = b,
    commentstyle = \color{codeblue},
    keywordstyle = \color{codekw},
}

\definecolor{green}{HTML}{009000}
\definecolor{red}{HTML}{ea4335}

\title{\our: Scaling Robotic Manipulation Foundation Models with 3D-Grounded Latent Actions}

\author[* 1, 2, 3]{Haoyu Zhao}
\author[* 1, 6]{Xingyue Zhao}
\author[* 1]{Zhian Su}
\author[* 1]{Bohan Hou}
\author[* 1]{Yinzhou Tang}
\author[2, 3]{\\ Fengyi Wang}
\author[1]{Kehan Li}
\author[5]{Yuming Jiang}
\author[1, 4]{Zhenyu Zeng}
\author[1, 4]{\\Mingxiu Chen}
\author[1, 4]{Deli Zhao}
\author[\dagger 1, 4]{Siteng Huang}
\author[1, 4]{Xin Li}
\author[\dagger 2, 3]{Zhongyu Li}

\affiliation[1]{DAMO Academy, Alibaba Group,}
\affiliation[2]{Hong Kong Embodied AI Lab,}
\affiliation[3]{CUHK,}
\affiliation[4]{Hupan Lab,\\}
\affiliation[5]{Alibaba Group,}
\affiliation[6]{PUMC}

\contribution[*]{Core contributors}
\contribution[\dagger]{Corresponding author}

\abstract{
Scaling vision-language-action (VLA) models is constrained by scarce action-labeled demonstrations and incompatible control spaces across robot embodiments. We present \textbf{\our}, a framework that turns heterogeneous interaction videos into scalable supervision for robot policy pretraining. At its core, \textbf{\lam} learns continuous, 3D-grounded latent actions from visual transitions without requiring native robot controls. By combining residual future-feature prediction, camera-compensated 3D-flow supervision, and camera-motion decoupling, \lam encourages representations that capture control-relevant motion rather than appearance changes alone. We apply \lam to 30.1K hours of human, robot, and simulation videos, producing 745.52M latent-action descriptors. \our then learns to predict latent-action chunks from visual observations and language through flow matching, before adapting its pretrained backbone and action expert to embodiment-specific robot actions. The resulting 4B policy achieves 98.8\% success on LIBERO, 84.7\% on LIBERO-Plus without benchmark-specific adaptation, 51.9\% average success across five VLABench tracks, and 88.6\% average success on the RoboTwin 2.0 bimanual benchmark.
It also achieves the highest observed success rates among the compared methods on six real-world manipulation tasks across dexterous-hand and mobile-gripper platforms. Together with representation ablations and data-scaling results, these findings demonstrate the potential of latent-action supervision for scaling robot policy learning beyond native action-labeled data.

\begin{center}
    \renewcommand{\arraystretch}{1.2}
    \begin{tabular}{ll}
        \homepage  & \url{https://alibaba-damo-academy.github.io/RynnVLA-Latent.github.io} \\
        \github  & \url{https://github.com/alibaba-damo-academy/RynnVLA-Latent}\\
        \huggingface & \url{https://huggingface.co/collections/Alibaba-DAMO-Academy/rynnvla-latent} \\
        \modelscope  & \url{https://www.modelscope.cn/collections/DAMO_Academy/RynnVLA-Latent}\\
    \end{tabular}
\end{center}

}

\date{\today}

\begin{document}
\thispagestyle{firstheader}
\maketitle
\pagestyle{empty}

\section{Introduction}
Large-scale pretraining has enabled broad capabilities in language and vision-language models~\citep{brown2020language,hoffmann2022training,kaplan2020scaling,achiam2023gpt,chen2022pali,bai2025qwen3,li2026rynnbrain}. Recent vision-language-action (VLA) and world-action models extend this approach to robot control~\citep{black2024pi_0,intelligence2025pi05,bjorck2025gr00t,wu2026pragmatic,li2026causal,ye2026world,yuan2026fast,bi2026motus,zhao2026towards}. Scaling policy learning, however, requires supervision that connects visual observations and task instructions to actions across diverse embodiments~\cite{zhao2026high,zhao2025physsplat}.

Robot demonstrations are expensive to collect, and their native controls differ in dimensions, coordinate frames, and semantics~\citep{zhao2026rynnworld,zhao2026rynnworld_teleop}. Human and web videos provide much broader interaction coverage but lack corresponding robot commands. These sources cannot simply be pooled under one native-action prediction objective. A shared supervision interface is needed to exploit their visual transitions before adapting a policy to a particular robot.

Latent actions provide such an interface by encoding changes between observations without native control labels~\citep{ye2025latent,chen2025moto,bu2025univla,gao2026dreamdojo,gao2025adaworld,schmidt2024learning,chen2024igor}. Existing approaches demonstrate the value of transition reconstruction and feature-space prediction, but visual change can include camera motion, static appearance, and effects unrelated to the agent's control. A representation that reconstructs images well therefore need not be equally useful for predicting low-level manipulation actions.

We address this through geometric supervision and explicit camera-motion separation. We introduce \textbf{\lam}, which encodes a frame pair into motion tokens, a global action latent, and a camera latent. Residual future-feature prediction preserves transition information through the bottleneck, while camera-compensated 3D-flow pseudo-labels and contrastive feature warping encourage motion-sensitive representations. Camera-pose prediction and gradient reversal further encourage separation of ego-motion from manipulation-related cues. These objectives provide geometric grounding without requiring manually annotated robot actions; they encourage camera decoupling rather than guaranteeing exact viewpoint invariance or physical identifiability.

Using \lam as an offline labeler, we construct shared latent-action supervision for 25 video sources spanning human interaction, real robots, and simulation. The corpus contains approximately 30.1K hours and 3.057B frames and yields 745.52M latent-action descriptors. We then pretrain \textbf{\our}, a Mixture-of-Transformers policy that couples a pretrained vision-language backbone with a narrow action expert. The policy predicts latent-action chunks from multi-view observations and language via flow matching, without native-action or proprioceptive supervision. In robot post-training, we reuse both the backbone and action-expert trunk, introduce proprioception, and retarget the output projection to native robot commands. This separates video-based motion supervision from embodiment-specific action alignment.

Our evaluation follows the same path from representation learning to robot deployment. On LARY, frozen-representation probes test whether the descriptors predict low-level control targets, with the full descriptor achieving average relative-action MSE $0.38$. After native-action post-training, the 4B policy obtains $98.8\%$ on LIBERO, $84.7\%$ on LIBERO-Plus, $51.9\%$ average success across five VLABench tracks, and $88.6\%$ average success under the RoboTwin 2.0 Native protocol. These benchmarks respectively assess task execution, perturbation robustness, semantic grounding, and bimanual manipulation. Six real-world tasks on stationary dexterous and mobile gripper platforms test the interface after adaptation to different controls. Data-scaling and component studies further characterize where the approach helps and where performance remains limited.

Our contributions are as follows:
\begin{itemize}[nosep, labelsep=0.6em, leftmargin=1.2em, itemindent=0em]
  \item \textbf{3D-grounded latent-action supervision.} We introduce RynnLAM, which combines residual feature prediction, camera-compensated 3D-flow supervision, and camera-motion decoupling to construct continuous transition representations without native robot-action labels.
  \item \textbf{Scalable VLA pretraining from heterogeneous video.} We develop RynnVLA-Latent, which learns shared latent-action targets from 30.1K hours of human, robot, and simulation videos, and transfers the pretrained policy to native robot controls through embodiment-specific post-training.
  \item \textbf{Representation and policy-level validation.} We demonstrate control predictability through frozen-representation probes, competitive manipulation performance across simulation and real-world platforms, and improvements with increasing latent-action pretraining data.
\end{itemize}

\section{Related Work}
\subsection{Latent Action Learning}

Latent action learning provides a way to exploit videos without native
action annotations by representing the transformations between observations.
LAPO~\citep{schmidt2024learning} and Genie~\citep{bruce2024genie}
demonstrate that latent action spaces can support policy learning and
controllable prediction from unlabeled videos.
LAPA~\citep{ye2025latent} learns discrete latent actions through
frame-transition reconstruction and uses them as VLA pretraining targets,
while Moto~\citep{chen2025moto} learns motion tokens for autoregressive
pretraining and transfers the resulting motion priors to robot control.
IGOR~\citep{chen2024igor} similarly compresses image-goal transitions into
shared action representations for learning across human and robot videos.
These approaches establish visual transitions as a scalable source of
action-like supervision beyond robot demonstrations.

Subsequent work investigates how to make such representations more useful
for control. UniVLA~\citep{bu2025univla} introduces language-guided,
task-centric latent actions in DINO feature space, while
villa-X~\citep{chen2026villa} augments visual reconstruction with a
proprioceptive forward-dynamics objective to ground latent actions in
robot states and actions.
Latent actions also serve as transferable conditioning signals for world
models: AdaWorld~\citep{gao2025adaworld} uses video-derived latent actions
for adaptable future prediction, and DreamDojo~\citep{gao2026dreamdojo}
learns continuous proxy actions from large-scale human videos.
Garrido et al.~\citep{garrido2026learning} further investigate constrained
continuous latent actions for representation-space world modeling on
in-the-wild videos.

Our focus is the geometric grounding of latent actions without robot-native
action or proprioceptive supervision.
\lam combines residual future-feature prediction, camera-compensated
dynamic 3D-flow supervision, and contrastive feature warping to learn a
factorized continuous transition representation.
A dedicated camera-pose pathway and an adversarial objective further
encourage separation of camera ego-motion from manipulation-relevant
motion. Rather than relying on continuity or feature-space prediction
alone, these complementary objectives encourage the latent representation
to capture physical scene changes and provide shared supervision across
human, simulated, and real-robot videos.

\subsection{Vision-Language-Action Models}

Vision-language-action (VLA) models transfer pretrained vision-language
knowledge to robot control.
OpenVLA~\citep{kim2024openvla} and FAST~\citep{pertsch2025fast}
explore action-token prediction, while
$\pi_0$~\citep{black2024pi_0}, GR00T N1~\citep{bjorck2025gr00t},
Xiaomi-Robotics-0~\citep{cai2026xiaomi}, and
LingBot-VLA~\citep{wu2026pragmatic} couple vision-language backbones
with continuous action generators.
Open X-Embodiment and RT-X~\citep{o2024open}, together with
$\pi_{0.5}$~\citep{intelligence2025pi05}, demonstrate the value of
cross-embodiment and heterogeneous training.
To handle incompatible action spaces,
UniAct~\citep{zheng2025universal} learns universal action representations
with embodiment-specific decoding,
X-VLA~\citep{zheng2026x} uses soft prompts and domain-specific
projections, and Qwen-RobotManip~\citep{yuan2026qwen} aligns canonical
state-action representations and camera-frame motion.
Human videos provide additional supervision through latent-action
labels in GR00T N1~\citep{bjorck2025gr00t} and reconstructed hand
motion in VITRA~\citep{li2025vitra}.

Our work focuses on constructing a shared supervision interface for scalable VLA pretraining, rather than aligning heterogeneous native control parameterizations. \our learns continuous \lam-generated latent-action chunks from human, robot, and simulation videos through flow matching.
The same descriptors also condition cross-embodiment video generation in the companion RynnWorld-Latent model.
Native robot actions and proprioception are introduced only during embodiment-specific post-training, allowing large-scale video-based pretraining to be separated from executable control alignment.

\begin{figure*}[t]
    \centering
    \includegraphics[width=\textwidth]{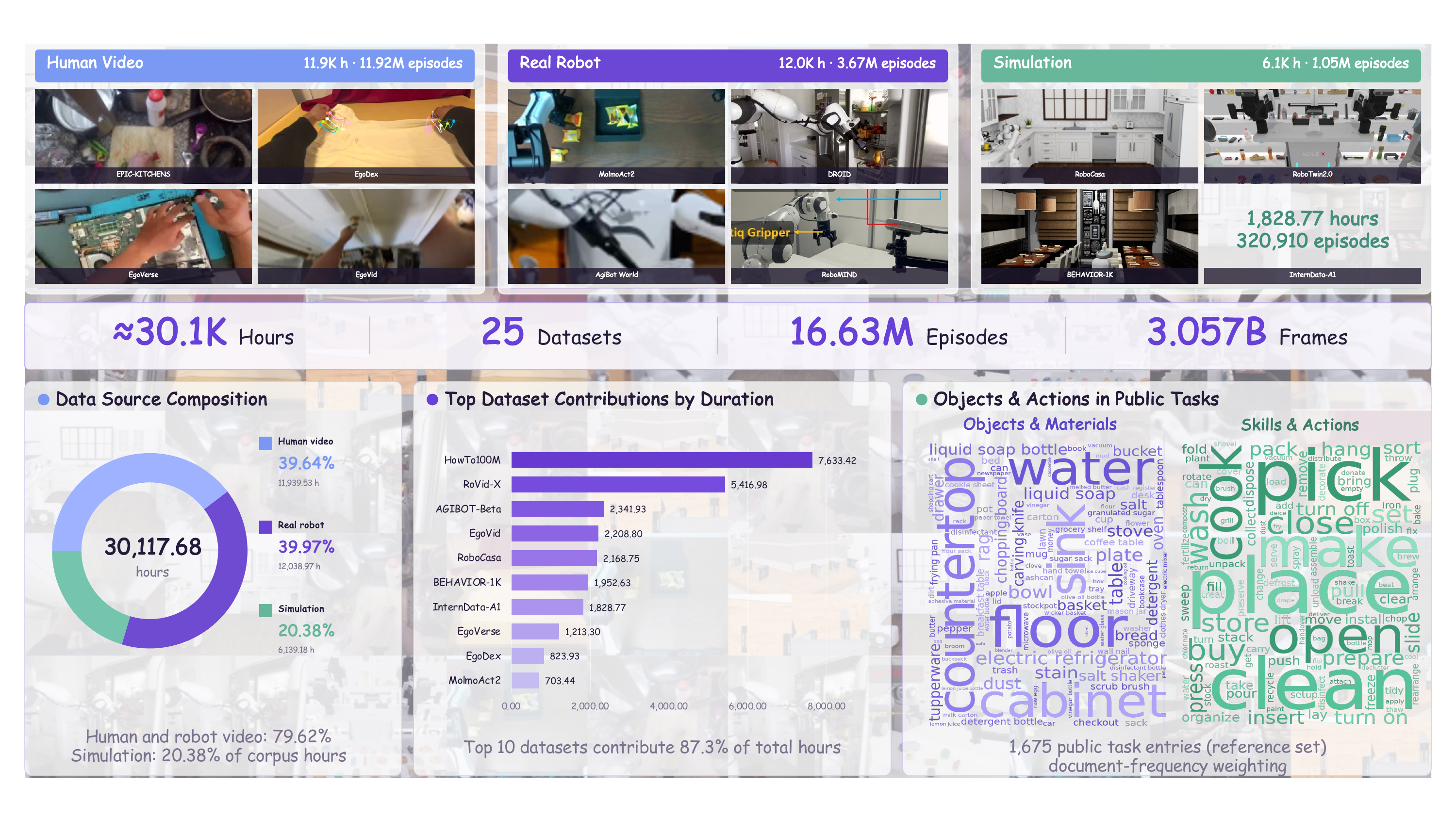}
    \caption{
    \textbf{Overview of the \our pretraining corpus.}
    We aggregate approximately 30.1K hours from 25 datasets, comprising 16.63M episodes and 3.057B frames. Human video and real-robot video each contribute approximately 40\% of the duration, with simulation accounting for the remaining 20\%. The figure summarizes source composition, dataset contributions by duration, and object and action vocabulary from public tasks.
    }
    \label{fig:pretraining_data}
\end{figure*}

\section{\our}

\label{sec:method}

A central obstacle to scaling robotic manipulation is that the most abundant sources of manipulation knowledge---human videos, web videos, simulation, and heterogeneous robot datasets---do not share a common action representation. We address this problem by learning a shared latent-action space supervised by visual transitions and geometric pseudo-labels, without requiring identical native control spaces. This latent space serves both as an automatic annotation for large-scale unlabeled video and as an intermediate control representation that can later be aligned with concrete robot actions. Our framework consists of three components. We first assemble and preprocess the multi-embodiment video corpus (Sec.~\ref{sec:data}). We then train \lam (Sec.~\ref{sec:lam}) with future-feature prediction, dynamic 3D-flow supervision, and camera-motion decoupling, and use its encoder to label the corpus offline. Finally, we train \our (Sec.~\ref{sec:our_vla}), a vision-language-action policy that learns to generate these latent actions from visual observations and language via flow matching. In post-training, the same flow-matching objective is retargeted from latent actions to the target robot's action space, so that costly action-labeled demonstrations are required only for the final embodiment alignment.

\begin{figure}
  \centering
    \includegraphics[width=0.9\linewidth]{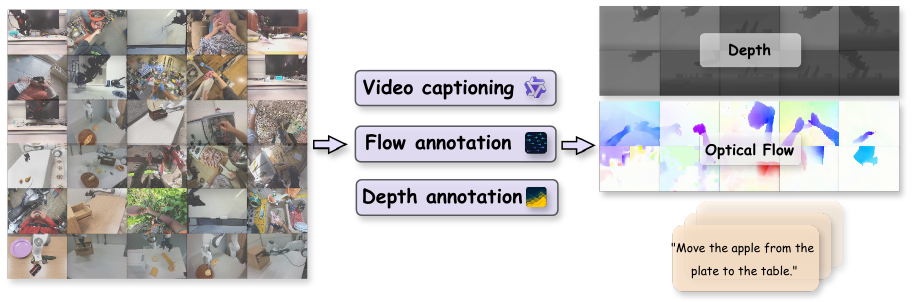}
  \caption{\textbf{Data curation pipeline.} The video data is collected from diverse sources and partitioned into short clips during data preprocessing. Each clip undergoes a multi-modal annotation process: (1) Video Captioning: Qwen3-VL~\citep{bai2025qwen3} generates detailed natural language descriptions of the video content; (2) Optical Flow Estimation: DPFlow~\citep{morimitsu2025dpflow} computes dense per-frame motion fields; (3) Depth Estimation: Depth Anything 3~\citep{lin2025depth} produces monocular depth predictions.}
  \label{fig:data-generation}
\end{figure}

\subsection{Pretraining Data}
\label{sec:data}

We collect a multi-embodiment corpus from \textbf{25} video sources, covering real-world robot demonstrations, simulated rollouts, and human interaction video (egocentric and web sources) across diverse embodiments (Fig.~\ref{fig:pretraining_data}). Every source is processed by the same pipeline without the need for action labels (Fig.~\ref{fig:data-generation}).

We process the entire dataset in parallel across $250$ shards using \lam. Specifically, we extract latent actions from consecutive frame pairs sampled every $4$ frames, i.e., pairs
$(0,4),(4,8),(8,12),\dots$, yielding one latent-action descriptor per adjacent pair in the four-frame sampling grid.

The source corpus contains approximately \textbf{30.1K hours}, \textbf{16.63M episodes}, and \textbf{3.057B frames} (Tab.~\ref{tab:pretraining-data-statistics}). Offline labeling produces \textbf{745.52M latent actions}. For policy pretraining, we form chunks containing $C{=}6$ latent-action targets from the offline annotation index. The RGB-frame stride used to extract a descriptor and the temporal spacing used to form a policy chunk are distinct indexing choices. The resulting policy training index contains \textbf{121.06M samples}, corresponding to approximately \textbf{946K optimizer steps} per epoch at a global batch size of $128$.

\begin{table*}[t]
  \centering
  \caption{\textbf{Dataset-level statistics} of the multi-source video corpus: all 25 sources listed below. Hours, episode counts, and frame counts summarize the source data. FPS denotes the frame rate declared in the source metadata rather than a uniform sampling rate, and can vary within a dataset; hours and frame counts are therefore not always consistent with this column.}
  \label{tab:pretraining-data-statistics}
  \small
  \setlength{\tabcolsep}{6pt}
  \renewcommand{\arraystretch}{1.08}
  
  \sisetup{
      group-separator={,}, 
      group-minimum-digits=4,
      table-number-alignment=center, 
      detect-weight=true, 
      mode=text
  }

  \begin{adjustbox}{max width=\linewidth}
    \begin{tabular}{@{}lc
          S[table-format=5.2, round-mode=places, round-precision=2]
          S[table-format=8.0, round-mode=none]
          S[table-format=5.3, round-mode=places, round-precision=3]@{}}
      \toprule
      \textbf{Dataset} & \textbf{FPS} & \textbf{Hours} & \textbf{Episodes} & {\textbf{Frames (M)}} \\
      \midrule
      
      \multicolumn{5}{@{}l}{\textit{Robot data}} \\
      \addlinespace[2pt]
      RoVid-X~\citep{deng2026rethinking} & 30 & 5416.98 & 2820534 & 585.034 \\
      AGIBOT-Beta~\citep{bu2025agibot} & 30 & 2341.93 & 141959 & 252.929 \\
      MolmoAct2~\citep{fang2026molmoact2} & 30 & 703.44 & 32239 & 75.972 \\
      RoboCOIN~\citep{wu2025robocoin} & 30 & 702.50 & 102017 & 76.619 \\
      DROID~\citep{khazatsky2024droid} & 15 & 507.25 & 94031 & 27.391 \\
      AgiBotWorld2026~\citep{bu2025agibot} & 30 & 443.24 & 20723 & 47.870 \\
      RoboMIND2.0~\citep{hou2025robomind} & 26 & 424.78 & 76745 & 34.124 \\
      Table30v2~\citep{yakefu2025robochallenge} & 30 & 423.75 & 32939 & 45.765 \\
      Galaxea~\citep{jiang2025galaxea} & 15 & 400.33 & 17040 & 21.688 \\
      Table30~\citep{yakefu2025robochallenge} & 30 & 288.40 & 25606 & 31.147 \\
      RoboMIND~\citep{wu2024robomind} & 20 & 155.16 & 43611 & 11.172 \\
      DreamDojo~\citep{gao2026dreamdojo} & 20 & 87.90 & 22185 & 6.329 \\
      OpenX-Embodiment~\citep{o2024open} & 30 & 76.16 & 213497 & 3.011 \\
      BridgeV2~\citep{walke2023bridgedata} & 5 & 37.58 & 16218 & 0.676 \\
      RDT-1B~\citep{liu2025rdt} & 30 & 29.57 & 6089 & 3.193 \\
      \addlinespace[1pt]
      \rowcolor{gray!10} \textbf{Subtotal} & -- & \bfseries 12038.97 & \bfseries 3665433 & \bfseries 1222.920 \\
      
      \midrule
      \multicolumn{5}{@{}l}{\textit{Robot data (simulation)}} \\
      \addlinespace[2pt]
      RoboCasa~\citep{nasiriany2024robocasa} & 20 & 2168.75 & 582567 & 156.150 \\
      BEHAVIOR-1K~\citep{li2023behavior} & 30 & 1952.63 & 20000 & 210.884 \\
      InternData-A1~\citep{tian2026interndata} & 30 & 1828.77 & 320910 & 197.508 \\
      RoboTwin 2.0~\citep{chen2025robotwin} & 30 & 189.03 & 125501 & 20.415 \\
      \addlinespace[1pt]
      \rowcolor{gray!10} \textbf{Subtotal} & -- & \bfseries 6139.18 & \bfseries 1048978 & \bfseries 584.957 \\
      
      \midrule
      \multicolumn{5}{@{}l}{\textit{Human interaction video}} \\
      \addlinespace[2pt]
      HowTo100M~\citep{miech2019howto100m} & 30 & 7633.42 & 9493227 & 784.274 \\
      EgoVid~\citep{wang2026egovid} & 30 & 2208.80 & 1963777 & 238.550 \\
      EgoVerse~\citep{punamiya2026egoverse} & 30 & 1213.30 & 57621 & 131.036 \\
      EgoDex~\citep{hoque2025egodex} & 30 & 823.93 & 338181 & 88.984 \\
      SthSthV2~\citep{goyal2017something} & 12 & 33.34 & 32968 & 1.440 \\
      EPIC\_KITCHENS~\citep{damen2020epic} & 60 & 26.74 & 31875 & 5.288 \\
      \addlinespace[1pt]
      \rowcolor{gray!10} \textbf{Subtotal} & -- & \bfseries 11939.53 & \bfseries 11917649 & \bfseries 1249.572 \\
      
      \midrule
      \addlinespace[1pt]
      \rowcolor{tablepurple} \textbf{Total corpus (25 datasets)} & -- & \bfseries 30117.68 & \bfseries 16632060 & \bfseries 3057.449 \\
      \bottomrule
    \end{tabular}
  \end{adjustbox}
\end{table*}

The corpus varies widely in embodiment, duration, and episode length. HowTo100M alone contributes more episodes than all real-robot sources combined, even though the human and robot groups have similar total duration. Shared latent-action labels make these sources compatible with one prediction objective, while source-weighted sampling controls their relative contribution during policy pretraining (Sec.~\ref{sec:impl_details}).

\begin{figure}[t]
    \centering
    \includegraphics[width=\linewidth]{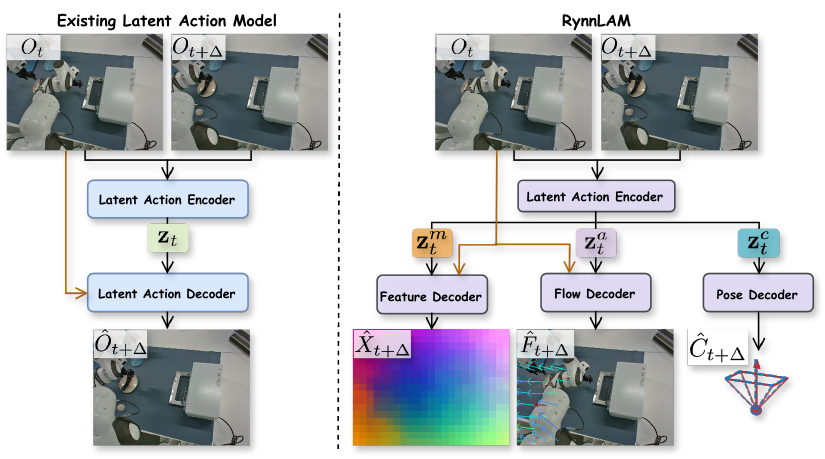}
    \caption{
    \textbf{Architecture of \lam and comparison with existing methods.} 
    \textbf{(a)} Conventional latent action models compress a frame transition into a single latent representation and reconstruct the future observation conditioned on the source frame. 
    \textbf{(b)} \textbf{\lam} factorizes transition information into motion tokens $z_t^m$, an action latent $z_t^a$, and a camera latent $z_t^c$. The three components are trained through complementary objectives: feature-space reconstruction, 3D dynamic-flow prediction, and relative camera-pose estimation, respectively. This factorization encourages the action representation to capture physical scene dynamics while encouraging camera ego-motion to be represented separately.
    }
    \label{fig:rynnlam}
\end{figure}

\subsection{\lam}
\label{sec:lam}

To enable action alignment across diverse embodiments, including heterogeneous robots and humans, we adopt latent actions as shared proxy representations of motion between a video frame pair. Several prior works learn a quantized codebook of latent actions~\citep{ye2025latent,chen2025moto}. Transition reconstruction provides a shared learning target, but does not explicitly distinguish object motion from camera-induced visual change. Our goal is to add geometric supervision that makes the representation more useful for control. To this end, \lam requires the latent representation to support three complementary objectives: predicting \textbf{future visual features}, recovering \textbf{dynamic 3D motion}, and disentangling \textbf{camera ego-motion from scene motion}.

\lam is built upon Depth-Anything-3~\citep{lin2025depth}; Fig.~\ref{fig:rynnlam} contrasts this factorized design with conventional latent-action models that reconstruct the future observation. Given a pair of RGB frames $(\mathbf{O}_t, \mathbf{O}_{t+\Delta})$, the encoder outputs a factorized continuous latent action
\begin{equation}
  \label{eq:latentaction}
  \mathbf{z}_t = [\mathbf{z}_t^{m}; \mathbf{z}_t^{a}; \mathbf{z}_t^{c}],
\end{equation}
where $\mathbf{z}_t^{m} \in \mathbb{R}^{512}$ collects $8$ local motion tokens ($8\times64$), $\mathbf{z}_t^{a} \in \mathbb{R}^{64}$ is a global action latent, and $\mathbf{z}_t^{c} \in \mathbb{R}^{32}$ is a dedicated camera latent. The encoder first computes a gated temporal difference between backbone features of the two frames, making the visual change explicit while suppressing irrelevant static appearance. Learned readout tokens summarize the global transition into $\mathbf{z}_t^{a}$ and $\mathbf{z}_t^{c}$, while learned motion queries attend to the target-frame features and produce compact motion tokens $\mathbf{z}_t^{m}$. This design allows the latent action to encode both global transition cues and local motion information without relying on explicit action labels.

The latent action is trained with self-supervised objectives that require no manual annotation. We minimize a combined objective of the form:
\begin{equation}
\label{eq:lam}
\begin{aligned}
\mathcal{L}_{\mathrm{LAM}}
&= \lambda_{\mathrm{rec}}\, \mathcal{L}_{\mathrm{rec}}
 + \lambda_{\mathrm{flow}}\, \mathcal{L}_{\mathrm{flow}}
 + \lambda_{\mathrm{pose}}\, \mathcal{L}_{\mathrm{pose}} \\
&\quad
 + \lambda_{\mathrm{feat}}\, \mathcal{L}_{\mathrm{feat}}
 + \lambda_{\mathrm{grl}}\, \mathcal{L}_{\mathrm{grl}},
\end{aligned}
\end{equation}
with $\lambda_{\mathrm{rec}}{=}\lambda_{\mathrm{flow}}{=}\lambda_{\mathrm{pose}}{=}5$, $\lambda_{\mathrm{feat}}{=}3$, and $\lambda_{\mathrm{grl}}{=}2$.

The reconstruction term $\mathcal{L}_{\mathrm{rec}}$ predicts the residual target-feature change $\mathbf{X}_{t+\Delta}-\mathbf{X}_t$ from the source features and the motion tokens $\mathbf{z}_t^{m}$. Although $\mathbf{z}_t^{m}$ is produced from target-frame features, it enters the reconstruction decoder as a compact latent bottleneck and serves as the decoder's only cross-frame input.
The $8\times64$-dimensional motion-token bottleneck compresses target-frame information, encouraging the representation to preserve compact transition cues rather than directly reproducing the full target feature map. Together with residual future-feature prediction, this bottleneck encourages the motion tokens to encode changes between observations while reducing reliance on static appearance. The resulting compact transition representation can then serve as part of the full latent-action descriptor predicted by the downstream policy without access to future frames at inference time.
Predicting a zero residual incurs an error wherever the target features differ from the source features.
Normalizing by the magnitude of the target change emphasizes transition reconstruction relative to its scale. Concretely, we use a normalized residual error,
\begin{equation}
\mathcal{L}_{\mathrm{rec}}
=\frac{\left\lVert(\hat{\mathbf{X}}_{t+\Delta} - \mathbf{X}_t)-(\mathbf{X}_{t+\Delta} -\mathbf{X}_t)\right\rVert^{2}}{\left\lVert\mathbf{X}_{t+\Delta} - \mathbf{X}_t\right\rVert^{2} + \epsilon_{0}},
\end{equation}
which supervises the predicted change rather than absolute feature appearance, and $\epsilon_{0}$ is a small constant that stabilizes the denominator when the two frames are nearly identical.

The flow term $\mathcal{L}_{\mathrm{flow}}$ predicts dynamic 3D scene flow in the source camera frame, using source features, monocular depth, and the action latent. Pseudo 3D-flow supervision is obtained from estimated depth, camera poses, and off-the-shelf optical flow. 
After compensating for estimated camera motion, static regions should have approximately zero dynamic flow, encouraging the latent to capture object-level physical motion rather than 2D pixel shifts. To further reduce reliance on static source appearance, we additionally mask a portion of source patch tokens during flow decoding.

The feature term $\mathcal{L}_{\mathrm{feat}}$ is contrastive rather than a plain consistency loss. Given the flow-warped source features $\tilde{\mathbf{X}}_t$, we require them to be closer to the target features $\mathbf{X}_{t+\Delta}$ than to the un-warped source features $\mathbf{X}_t$,
\begin{equation}
\mathcal{L}_{\mathrm{feat}}
=
\mathbb{E}
\left[
w_t\,
\max\!\Bigl(0,\; \mu
- s\bigl(\tilde{\mathbf{X}}_t,\mathbf{X}_{t+\Delta}\bigr)
+ s\bigl(\tilde{\mathbf{X}}_t,\mathbf{X}_t\bigr)
\Bigr)
\right],
\end{equation}
where $s(\cdot,\cdot)$ denotes cosine similarity between the corresponding per-location feature vectors, $\mu$ is a margin, and $w_t$ weights each spatial location by its flow magnitude. Using the un-warped source as a negative discourages identity warping in regions where the source and target differ, while the motion-magnitude weighting focuses supervision on regions with non-trivial dynamics.

To separate viewpoint changes from manipulation-relevant dynamics, we introduce explicit camera-pose decoupling. Let $g_{\mathrm{pose}}(\cdot)$ denote a pose decoder that maps a latent to the relative $6$-DoF camera pose $\mathbf{p}_t^{\mathrm{cam}}$ between the two frames. The camera latent $\mathbf{z}_t^{c}$ is supervised to decode this pose, $\mathcal{L}_{\mathrm{pose}}=\bigl\lVert g_{\mathrm{pose}}(\mathbf{z}_t^{c})-\mathbf{p}_t^{\mathrm{cam}}\bigr\rVert^{2}$, while a gradient-reversal objective drives the same decoder to fail on the action latent $\mathbf{z}_t^{a}$,
\begin{equation}
\mathcal{L}_{\mathrm{grl}}
=
\left\lVert
g_{\mathrm{pose}}\bigl(\mathrm{GRL}(\mathbf{z}_t^{a})\bigr)
-
\mathbf{p}_t^{\mathrm{cam}}
\right\rVert^{2},
\end{equation}
where $\mathrm{GRL}(\cdot)$ forwards its input unchanged and negates the gradient during backpropagation. As a result, $\mathbf{z}_t^{a}$ is encouraged to retain interaction- and motion-relevant cues, while ego-motion is routed into the separate camera latent.

For offline labeling, only the encoder is required. It processes multi-embodiment frame pairs and extracts the concatenated descriptor $[\mathbf{z}^{m};\mathbf{z}^{a};\mathbf{z}^{c}]$. Camera-pose decoupling acts specifically on the global action latent; the full descriptor retains the camera latent and is therefore not assumed to be camera-invariant. The motion tokens and global latents provide complementary transition information, with the full 608-dimensional descriptor achieving better control predictability than any individual component in our representation probes (Appendix~\ref{app:extended_ablations_lam}). We therefore use the complete descriptor as the shared supervision target for policy pretraining, without requiring manual alignment of native control spaces. The decoders are retained only for qualitative visualization.

\begin{figure*}[t]
    \centering
    \includegraphics[width=\textwidth]{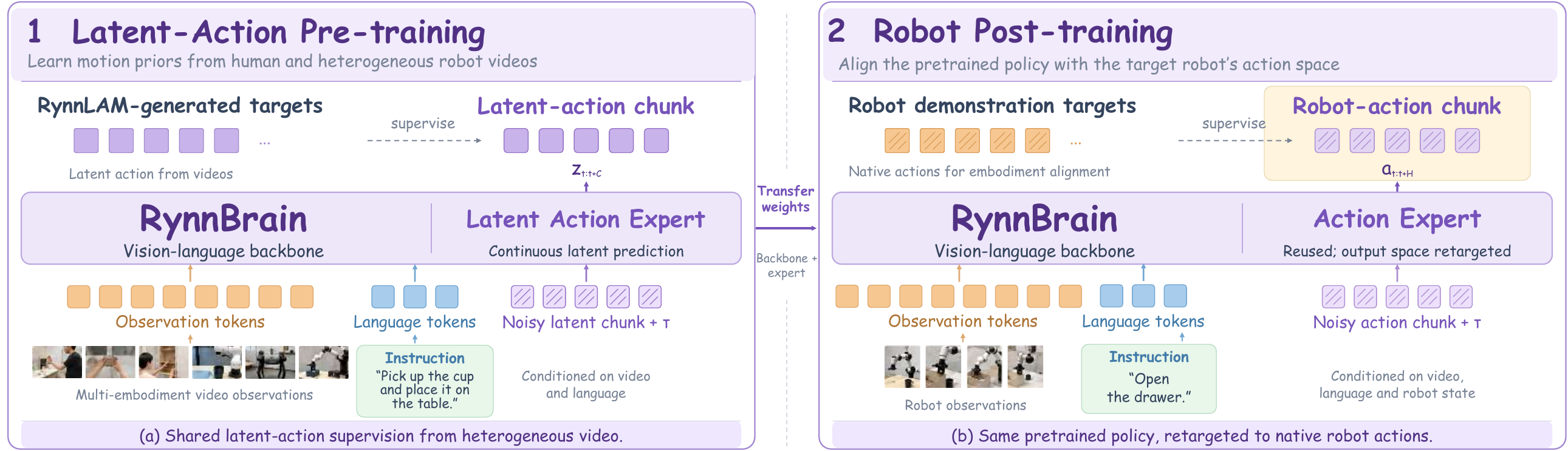}
    \caption{
    \textbf{Two-stage training pipeline of \our.}
    \textbf{(a) Latent-action pretraining.} \lam provides shared transition descriptors for human and heterogeneous robot videos. The policy predicts these descriptors from observations and language.
    \textbf{(b) Robot post-training.} Both the pretrained RynnBrain backbone and action-expert trunk are reused. The output projection is retargeted to native robot actions, and proprioceptive state is introduced. Post-training adapts the pretrained policy to the target control space without adding a separate latent-to-action decoder.
    }
    \label{fig:rynnvla_pipeline}
\end{figure*}

\subsection{\our}
\label{sec:our_vla}

The latent actions produced by \lam provide an embodiment-agnostic interface between video understanding and control; Fig.~\ref{fig:rynnvla_pipeline} summarizes the resulting training pipeline. Because the same encoder maps heterogeneous videos into a common descriptor layout, human, robot, and simulation videos can contribute to one policy objective without reconciling their native action conventions. Building on this interface, \our separates the learning of generalizable manipulation dynamics from embodiment-specific action alignment. During pre-training, the policy learns to generate latent actions from raw observations and language; during post-training, the same flow-matching objective is retargeted from latent actions to the target robot's action space. This decoupling allows the model to scale on unlabeled multi-embodiment videos, while requiring costly action-labeled demonstrations only for the final embodiment alignment.

\paragraph{Architecture.}
\our adopts a Mixture-of-Transformers (MoT) architecture~\citep{liang2024mixture} that couples a pretrained vision-language model with a narrow action expert. The VLM backbone, RynnBrain~\citep{li2026rynnbrain}, jointly encodes multi-view images and the language instruction into a causal prefix. The action expert is a randomly initialized transformer stream with the same depth as the VLM but a smaller hidden width. At each layer, tokens from the two streams are jointly attended, while their output projections and feed-forward pathways remain separate. The VLM prefix remains causal and cannot attend to the action suffix, whereas the action suffix is bidirectional and attends to the full VLM prefix. This design allows the action expert to exploit rich visual-language context without duplicating the full VLM capacity, and permits action-prediction gradients to refine the VLM representation. We instantiate \our with 2B and 4B VLM backbones while keeping the action expert narrow. The 2B and 4B names refer to the backbone variants; total policy parameter counts, including the action expert, are reported in Tab.~\ref{tab:model_config}.

\paragraph{Pre-training: flow matching on latent actions.}
We use half-open chunk notation: $\mathbf{z}_{t:t+C}=(\mathbf{z}_t,\ldots,\allowbreak\mathbf{z}_{t+C-1})$ contains exactly $C$ latent actions. For each sample $i$ and each available view $v\in\mathcal{V}_i$, \lam provides a latent-action chunk $\mathbf{z}^{(v)}_{t:t+C}$, where $\mathcal{V}_i$ denotes the set of views present in that sample. The pre-training objective is to learn a conditional velocity field that generates these chunks from Gaussian noise. Given
\begin{equation}
\mathbf{z}^{\tau,(v)}_{t:t+C}
=
\tau\,\boldsymbol{\epsilon}^{(v)}
+
(1-\tau)\,\mathbf{z}^{(v)}_{t:t+C},
\qquad
\boldsymbol{\epsilon}^{(v)}\sim\mathcal{N}(\mathbf{0},\mathbf{I}),
\end{equation}
where $\tau{=}1$ corresponds to pure noise and $\tau{=}0$ to the clean latent, the expert predicts the derivative of this interpolation with respect to $\tau$,
$\boldsymbol{\epsilon}^{(v)}-\mathbf{z}^{(v)}_{t:t+C}$, conditioned on all present views, the language instruction, and the flow timestep $\tau$. The training objective is
\begin{equation}
\mathcal{L}_{\mathrm{FM}}
=
\mathbb{E}_{i,\tau,\boldsymbol{\epsilon}}
\frac{1}{|\mathcal{V}_i|}
\sum_{v\in\mathcal{V}_i}
\frac{1}{C d_z}
\left\|
\mathbf{v}^{(v)}_{\theta}
-
\big(
\boldsymbol{\epsilon}^{(v)}-\mathbf{z}^{(v)}_{t:t+C}
\big)
\right\|_2^2,
\end{equation}
where $d_z{=}608$ is the width of the concatenated latent action $\mathbf{z}_t$ in Eq.~\eqref{eq:latentaction}. By defining $\mathcal{V}_i$ as the set of present views, missing views contribute neither to attention nor to the loss. Timestep information is injected through adaptive normalization and input embeddings. We sample $u\sim\mathrm{Beta}(1.5,1)$ and linearly map $u$ to $[0.001,0.999]$, biasing training toward noisier timesteps. Both the VLM and the action expert are updated. This stage requires neither robot actions nor proprioception; its supervision is available for any video from which \lam can extract latent actions.

\paragraph{Multi-view conditioning.}
Embodiments differ not only in action spaces but also in camera configurations. To avoid hard-coding a fixed set of views, \our organizes latent-action generation using canonical camera slots. For each present view, its VLM patch embeddings are pooled into a slot token and prepended to the action expert. Empty slots are masked in attention and in the loss. This slot-based interface accommodates samples with different available views without supervising absent outputs. Camera configurations are mapped to the canonical slots; the interface alone does not guarantee invariance to an unseen viewpoint.

\begin{table}[t]
  \caption{
    \textbf{Model configurations for the scaling variants of \our}. Total parameters denote the total number of model parameters.
  }
  \label{tab:model_config}
  \centering
  \setlength{\tabcolsep}{4.5pt}
  \begin{tabular}{l c cc cc c}
    \toprule
    \multirow{2}{*}{\textbf{Method}} &
    \multirow{2}{*}{\textbf{Layers}} &
    \multicolumn{2}{c}{\textbf{VLM}} &
    \multicolumn{2}{c}{\textbf{Action Expert}} &
    \multirow{2}{*}{\textbf{\begin{tabular}{@{}c@{}}Total Params\end{tabular}}} \\
    \cmidrule(lr){3-4} \cmidrule(lr){5-6}
    & &
    {\small\begin{tabular}{@{}c@{}}Hidden Size\end{tabular}} &
    {\small\begin{tabular}{@{}c@{}}Params\end{tabular}} &
    {\small\begin{tabular}{@{}c@{}}Hidden Size\end{tabular}} &
    {\small\begin{tabular}{@{}c@{}}Params\end{tabular}} & \\
    \midrule
    \textbf{{\our}-2B} & 28 & 2048 & 2.13\,B & 768 & 0.38\,B & \textbf{2.51\,B} \\
    \textbf{{\our}-4B} & 36 & 2560 & 4.44\,B & 768 & 0.60\,B & \textbf{5.04\,B} \\
    \bottomrule
  \end{tabular}
\end{table}

\paragraph{Post-training: embodiment-specific action alignment.}
Pre-training yields a model that generates embodiment-agnostic latent actions but does not yet produce executable robot commands. In post-training, we retarget the same flow-matching objective from \lam latents to the target robot's action space and fine-tune the full model on demonstrations from that embodiment. Proprioceptive state is introduced only at this stage. No separate action decoder is added: the readout trunk pretrained during pre-training is retained and terminated by a thin newly initialized projection to the target action dimension, so the expert's pretrained capacity is reused rather than duplicated. We use a unified action layout and mask dimensions not used by the target embodiment, allowing one interface to support different robots.

\paragraph{Inference.}
At inference, \our initializes the output sequence from Gaussian noise and integrates the learned velocity field from $\tau{=}1$ toward $\tau{=}0$ using Euler steps. In pre-training this output is a latent-action chunk; after post-training it is directly the robot action chunk. The VLM prefix is computed once and reused across integration steps. The resulting action chunk is executed in full before replanning. Post-training updates the policy parameters and changes the prediction target to native actions while retaining this sampling procedure; execution requires no separate latent-to-action decoder.

\section{Experiments}

We evaluate the proposed framework from three complementary perspectives.
First, we assess the quality of the latent action representation itself, asking whether \lam learns a representation that is predictive of low-level control signals rather than merely encoding visual semantics or appearance change.
Second, we report completed downstream policy evaluations on simulation benchmarks covering manipulation success, robustness to perturbations, and language grounding. We additionally evaluate bimanual manipulation on RoboTwin 2.0 under the Native protocol.
Third, we evaluate post-trained policies on real-world dual-arm platforms with different end-effectors and mobility, testing applicability across these deployment settings after embodiment-specific adaptation.

\subsection{Implementation Details}
\label{sec:impl_details}

\noindent\textbf{\lam.}
\lam is built upon Depth-Anything-3~\citep{lin2025depth} and trained on video frame pairs without native-action labels, using visual-feature targets and geometric pseudo-labels. The model is optimized with the combined objective in Eq.~\eqref{eq:lam}, which supervises the latent action through residual feature prediction, dynamic 3D flow regression, contrastive warped-feature alignment, camera pose decoding, and gradient-reversal-based camera decoupling. 
Training proceeds in two stages: Stage 1 ($36$k steps, encoder frozen) and Stage 2 ($20$k steps, encoder unfrozen). The final $608$-dimensional descriptor comprises $8$ motion tokens ($512$-D), a global latent action ($64$-D), and a relative camera-pose latent ($32$-D). \lam is applied offline to the entire pretraining corpus (temporal stride $4$ frames) to generate the training index for \our.

\noindent\textbf{\our.}
\our adopts a Mixture-of-Transformers (MoT) architecture~\citep{liang2024mixture} consisting of a pretrained vision-language model and a narrow action expert. The VLM backbone is RynnBrain~\citep{li2026rynnbrain}. 
The action expert matches the VLM depth but uses a smaller width ($d_{\text{model}}{=}768$, $d_{\text{ff}}{=}2752$), interacting via per-layer joint attention. The observation interface supports $6$ view-conditioned slots, with observation frames sampled at $2$~fps, a maximum of $64$ frames, and a $4096$-token budget. Native end-effector rotations use a $6$-D representation during robot-action post-training.
During pretraining, \our is trained with flow matching to predict chunks of $6$ latent-action targets conditioned on multi-view observations and language instructions, where each latent is the $608$-dimensional \lam descriptor. Proprioception is excluded. Both the VLM and action expert are updated using AdamW ($\beta_1{=}0.9$, $\beta_2{=}0.95$, weight decay$=0$, gradient clipping$=1.0$). Sources are sampled with probability $p_d \propto W_d^{0.7}$, where $W_d$ is the configured weight for source $d$, to balance internet video and robot data. During post-training, the same flow-matching objective is retargeted from latent actions to the target robot's action space, and proprioceptive state is introduced. No separate decoder is added; the pretrained readout trunk is retained and terminated by a thin projection to the target action dimension.

\begin{figure}
  \centering
    \includegraphics[width=0.95\linewidth]{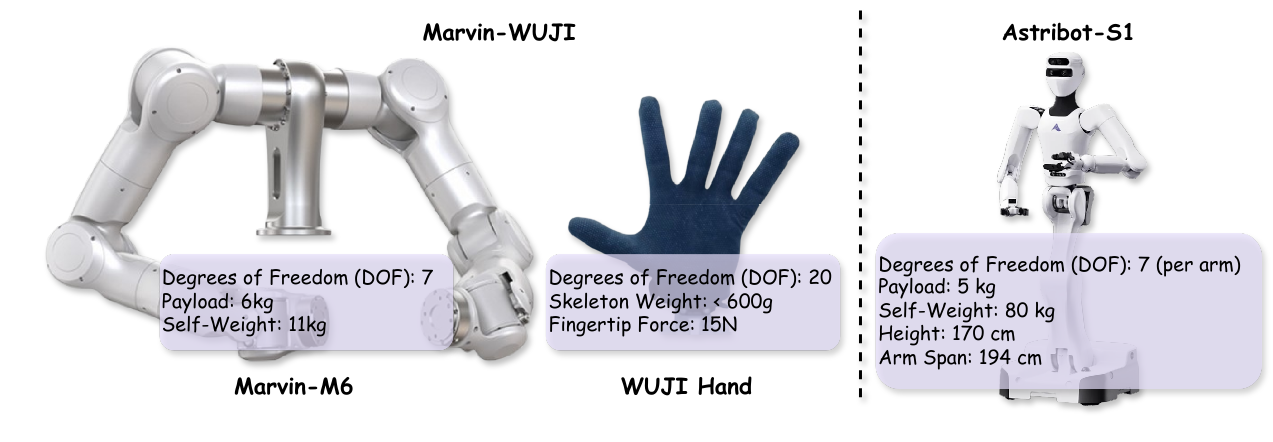}
  \caption{\textbf{Standardized Experimental Platforms.} Two dual-arm embodiments used for real-world evaluation: the stationary \textbf{Marvin-WUJI} (dual 7-DOF arms with 20-DOF dexterous hands) and the mobile \textbf{Astribot-S1} (dual 7-DOF arms with parallel grippers on a wheeled base).}
  \label{fig:vis-robot}
\end{figure}

\subsection{Real-Robot System Setup}

Our real-world experiments use two robotic platforms to evaluate deployment after embodiment-specific post-training (Fig.~\ref{fig:vis-robot}). The policy predicts native action chunks, which are passed to the robot control interface for execution. Communication uses LCM (Lightweight Communications and Marshalling). This evaluation measures closed-loop task success; inference speed is not an evaluation metric.

\noindent
\textbf{Marvin-WUJI.} Dual-Arm Dexterous Hand System. The first embodiment is built on the TIANJI M6 arms and WUJI dexterous hands. The hardware setup consists of dual Tianji 7-DOF robotic arms and dual Wuji 20-DOF dexterous hands, yielding a total of 54 degrees of freedom.

\noindent
\textbf{Astribot-S1.} Astribot-S1 Mobile Manipulator. The second embodiment utilizes the Astribot S1, an AI robot platform featuring a wheeled mobile base and dual 7-DOF robotic arms equipped with parallel grippers. Each arm provides 7 DOFs with a rated payload of 5\,kg (horizontal extension) and an end-effector repeatability of $\pm 0.1$\,mm. With a height of 170\,cm and an arm span of 194\,cm, Astribot-S1 complements the stationary dexterous platform with mobile bimanual manipulation and parallel grippers.

\subsection{Latent Action Representation Quality}
We evaluate the quality of the learned latent representation using the LARY benchmark~\citep{nie2026lary}. We focus on its low-level control regression track, which decodes end-effector trajectories from a frozen representation using a lightweight MLP. Since the official absolute targets can be partially predicted from the currently visible robot pose under short temporal strides, we additionally report a relative setting that predicts end-effector displacements from the first frame. This setting more directly measures whether the representation encodes motion rather than static pose. We compare with general semantic encoders, generative pixel encoders, embodied latent-action models, and general latent-action models, following~\citet{nie2026lary}.

\begin{table}[t]
  \caption{\textbf{Results and Ablation Study on the LARY benchmark~\citep{nie2026lary}}. Low-level control regression (MSE, lower is better). `\lam (Raw Features)' probes the latent representation produced directly by the \lam encoder without assembling the 608-dimensional descriptor, following the general-encoder protocol; the `\lam' rows probe the full 608-dimensional latent-action descriptor.}
  \label{tab:exp_lary}
  \centering
  \setlength{\tabcolsep}{6pt}
  \small
  \begin{tabular}{l ccccc}
    \toprule
    \textbf{Method} & \textbf{Avg.}  & \textbf{CALVIN} & \textbf{VLABench} & \textbf{RoboCOIN} & \textbf{AgiBotWorld-B} \\
    \midrule
    
    \multicolumn{6}{l}{\emph{General Encoders (Absolute Action)}} \\
    V-JEPA~2~\citep{assran2025vjepa2} & 0.25 & 0.27 & 0.07 & 0.32 & 0.33 \\
    DINOv3~\citep{simeoni2025dinov3}  & 0.19 & 0.22 & 0.06 & \textbf{0.22} & 0.24 \\
    Wan2.2~\citep{wan2025}   & 0.30 & 0.39 & 0.09 & 0.34 & 0.39 \\
    FLUX.2-dev~\citep{flux-2-2025}    & 0.35 & 0.25 & 0.04 & 0.47 & 0.62 \\
    \midrule
    \rowcolor{tablepurple}
    \textbf{\lam (Raw Features)} & \textbf{0.14} & \textbf{0.08} & \textbf{0.01} & 0.24 & \textbf{0.22} \\
    \midrule
    
    \multicolumn{6}{l}{\emph{Embodied LAMs (Absolute Action)}} \\
    LAPA~\citep{ye2025latent}  & 0.97 & 0.96 & 0.95 & 0.96 & 1.00 \\
    Moto~\citep{chen2025moto}  & 0.67 & \textbf{0.24} & 0.60 & 0.88 & 0.97 \\
    UniVLA~\citep{bu2025univla}   & 0.87 & 0.82 & 0.74 & 0.94 & 0.97 \\
    villa-X~\citep{chen2026villa}  & 0.87 & 0.86 & 0.72 & 0.94 & 0.97 \\
    DreamDojo~\citep{gao2026dreamdojo}  & 0.49 & 0.34 & 0.14 & 0.74 & 0.74  \\
    \midrule
    \rowcolor{tablepurple}
    \textbf{\lam} & \textbf{0.27} & 0.34 & \textbf{0.01} & \textbf{0.34} & \textbf{0.39} \\
    \midrule

    \multicolumn{6}{l}{\emph{Embodied LAMs (Relative Action)}} \\
    LAPA~\citep{ye2025latent}  & 0.93 & 0.72 & 1.00 & 1.01 & 0.98  \\
    Moto~\citep{chen2025moto}  & 0.77 & 0.65 & 0.64 & 0.91 & 0.89  \\
    DreamDojo~\citep{gao2026dreamdojo}  & 0.55 & \textbf{0.35} & 0.33 & 0.81 & 0.70  \\
    \midrule
    \rowcolor{tablepurple}
    \textbf{\lam} & \textbf{0.38} & 0.40 & \textbf{0.05} & \textbf{0.42} & \textbf{0.64} \\ 
    \midrule
    
    \multicolumn{6}{l}{\emph{Ablation Study (Absolute Action)}} \\
     \lam with frozen encoder & 0.37 & 0.41 & 0.10 & 0.43 & 0.52  \\
    \lam w/o 3D flow & 0.37 & 0.46 & 0.10 & 0.42 & 0.48  \\
    \lam w/o camera & 0.32 & 0.39 & 0.07 & 0.39 & 0.42 \\
    \lam w/o motion-token residual forcing & 0.37 & 0.45 & 0.09 & 0.43 & 0.52 \\
    \lam w/o GRL & 0.32 & 0.37 & 0.09 & 0.38 & 0.43 \\
    \midrule
    \rowcolor{tablepurple}
    \textbf{\lam} & \textbf{0.27} & \textbf{0.34} & \textbf{0.01} & \textbf{0.34} & \textbf{0.39} \\
    \midrule

    \multicolumn{6}{l}{\emph{Ablation Study (Relative Action)}} \\
    \lam with frozen encoder & 0.52 & 0.43 & 0.12 & 0.72 & 0.79 \\
    \lam w/o 3D flow & 0.50 & 0.46 & 0.10 & 0.70 & 0.75 \\
    \lam w/o camera & 0.47 & 0.43 & 0.09 & 0.65 & 0.69 \\
    \lam w/o motion-token residual forcing & 0.57 & 0.50 & 0.14 & 0.80 & 0.85  \\
    \lam w/o GRL & 0.49 & 0.43 & 0.07 & 0.72 & 0.73 \\
    \midrule
    \rowcolor{tablepurple}
    \textbf{\lam} & \textbf{0.38} & \textbf{0.40} & \textbf{0.05} & \textbf{0.42} & \textbf{0.64} \\

    \bottomrule
  \end{tabular}
\end{table}

\begin{figure}[t]
  \centering
  \includegraphics[width=0.95\linewidth]{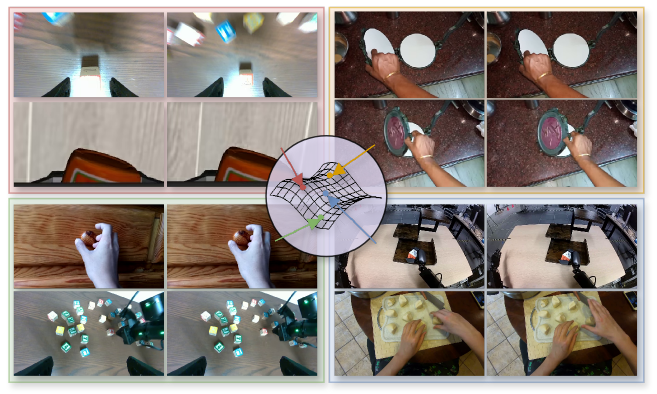}
  \caption{\textbf{Latent-action similarity visualization.}
    We retrieve the nearest neighbors of several query latent actions
    across the multi-embodiment corpus and display the corresponding
    video frames. The displayed high-similarity neighbors exhibit related motion patterns across different embodiments and visual domains, providing a qualitative illustration of cross-domain transition similarity.}
  \label{fig:vis-latent-similarity}
\end{figure}

\subsection{Simulation Benchmarks}
We report completed policy evaluations on LIBERO, LIBERO-Plus, and VLABench, covering manipulation success, robustness to perturbations, and semantic grounding. RoboTwin 2.0 Native evaluation additionally assesses bimanual manipulation. Detailed benchmark statistics and evaluation protocols are provided in Appendix~\ref{app:benchmark_details}.

\noindent\textbf{LIBERO and LIBERO-Plus.}
LIBERO~\citep{liu2023libero} evaluates language-conditioned single-arm manipulation across spatial, object, goal, and long-horizon task suites. LIBERO-Plus~\citep{fei25libero-plus} uses the same tasks but evaluates robustness under perturbations such as viewpoint, robot state, language, lighting, background, layout, and sensor noise. We use LIBERO to verify basic manipulation competence and LIBERO-Plus to measure robustness beyond in-distribution execution.

\noindent\textbf{VLABench.}
VLABench~\citep{zhang2025vlabench} evaluates semantic grounding and instruction understanding in manipulation. Its tasks require resolving target objects and intentions under implicit instructions, unseen categories, and visual variations. We report success rate together with the benchmark's progress and intention scores.

\noindent\textbf{RoboTwin 2.0.}
RoboTwin 2.0~\citep{chen2025robotwin} evaluates bimanual manipulation under clean and randomized scenes. We use the Native protocol, which includes both clean and randomized demonstrations during training. Tab.~\ref{tab:exp_robotwin} reports results for \our alongside the listed baselines.

\begin{table}[t]
  \caption{\textbf{Results on the LIBERO benchmark.}}
  \label{tab:exp_libero}
  \small
  \setlength{\tabcolsep}{10pt}
  \centering
  \begin{tabular}{l c c c c c}
    \toprule
    \textbf{Method} & \textbf{Spatial} & \textbf{Object} & \textbf{Goal} & \textbf{Long} & \textbf{Avg.} \\
    \midrule
    OpenVLA-OFT~\citep{kim2025fine}   & 97.6 & 98.4 & 97.9 & 94.5 & 97.1 \\
    $\pi_{0}$-FAST~\citep{pertsch2025fast}   & 96.4 & 96.8 & 88.6 & 60.2 & 85.5 \\
    $\pi_{0}$~\citep{black2024pi_0}  & 98.0 & 96.8 & 94.4 & 88.4 & 94.4 \\
    $\pi_{0.5}$~\citep{intelligence2025pi05}   & 98.8 & 98.2 & 98.0 & 92.4 & 96.9 \\
    GR00T-N1.7~\citep{bjorck2025gr00t}    & 97.7 & 98.5 & 97.5 & 94.4 & 97.0 \\
    InternVLA-M1~\citep{chen2025internvla} & 98.0 & 99.0 & 93.8 & 92.6 & 95.9 \\
    Motus~\citep{bi2026motus} & 96.8 & 99.8 & 96.6 & 97.6 & 97.7 \\
    Fast-WAM~\citep{yuan2026fast} & 98.2 & \textbf{100.0} & 97.0 & 95.2 & 97.6 \\
    Xiaomi-Robotics-0~\citep{cai2026xiaomi} & 98.8 & \textbf{100.0} & \textbf{98.8} & 97.2 & 98.7 \\
    LingBot-VA~\citep{li2026causal} & 98.5 & 99.6 & 97.2 & \textbf{98.5} & 98.5 \\
    \midrule
    \rowcolor{tablepurple} \textbf{\our-2B} &  98.6 & 99.6 & 97.0 & 
    97.8 & 98.3 \\
    \rowcolor{tablepurple} \textbf{\our-4B} & \textbf{99.4} & 99.8 & \textbf{98.8} & 97.0 & \textbf{98.8} \\
    \bottomrule
  \end{tabular}
\end{table}

\begin{table}
  \caption{\textbf{Results on LIBERO-Plus}, evaluated zero-shot with the LIBERO-finetuned checkpoint.}
  \label{tab:exp_liberoplus}
  \centering
  \setlength{\tabcolsep}{5pt}
  \small
  \begin{tabular}{l c c c c c c c c}
    \toprule
    \textbf{Method} & \textbf{Cam.} & \textbf{Robot} & \textbf{Lang.} & \textbf{Light} & \textbf{Bkg.} & \textbf{Noise} & \textbf{Layout} & \textbf{Total} \\
    \midrule
    OpenVLA~\citep{kim2024openvla}  & 0.8 & 3.5 & 23.0 & 8.1 & 34.8 & 15.2 & 28.5 & 15.6 \\
    OpenVLA-OFT~\citep{kim2025fine}  & 56.4 & 31.9 & 79.5 & 88.7 & 93.3 & 75.8 & 74.2 & 69.6 \\
    $\pi_{0}$~\citep{black2024pi_0}  & 13.8 & 6.0 & 58.8 & 85.0 & 81.4 & 79.0 & 68.9 & 53.6 \\
    $\pi_{0}$-FAST~\citep{pertsch2025fast}& 65.1 & 21.6 & 61.0 & 73.2 & 73.2 & 74.4 & 68.8 & 61.6 \\
    $\pi_{0.5}$~\citep{intelligence2025pi05}  & \textbf{78.4} & \textbf{73.6} & 80.8 & 96.2 & 94.1 & 89.0 & 84.5 & 84.4 \\
    StarVLA~\citep{community2026starvla} & 52.5 & 49.8 & 88.5 & 95.7 & \textbf{95.7} & 73.0 & 76.9 & 74.1 \\
    ABot-M0~\citep{yang2026abot} & 60.4 & 67.9 & 86.4 & 96.2 & 91.6 & 86.4 & 82.6 & 80.5 \\
    Cosmos Policy~\citep{kim2026cosmos}  & 75.8 & 63.3 & 81.7 & 96.5 & 88.9 & 92.7 & 82.2 & 82.2 \\
    \midrule
    \rowcolor{tablepurple} \textbf{\our-2B} & 76.4 & 59.3 & 88.3 & 96.2 & 94.6 & 89.6 & 83.2 & 82.9 \\
    \rowcolor{tablepurple} \textbf{\our-4B} & 72.9 & 64.0 & \textbf{91.7} & \textbf{97.1} & 95.4 & \textbf{93.2} & \textbf{85.2} & \textbf{84.7} \\
    \bottomrule
  \end{tabular}
\end{table}

\begin{table}[t]
  \caption{\textbf{RoboTwin 2.0 Native protocol.} Values are success rates in percent. Baseline values are taken from the respective reports. Avg. in our rows is the arithmetic mean of Clean and Randomized.}
  \label{tab:exp_robotwin}
  \centering
  \small
  \begin{tabular}{l c c c}
    \toprule
    \textbf{Method} & \textbf{Clean} & \textbf{Randomized} & \textbf{Avg.} \\
    \midrule
    $\pi_{0}$~\citep{black2024pi_0}    & 65.9 & 58.4 & 62.2 \\
    $\pi_{0.5}$~\citep{intelligence2025pi05}    & 82.7 & 76.8 & 79.8 \\
    X-VLA~\citep{zheng2026x} & 72.9 & 72.8 & 72.9 \\
    StarVLA-FAST~\citep{community2026starvla}   & 72.5 & 83.2 & 77.9 \\
    ABot-M0~\citep{yang2026abot} & 86.1  & 85.1 & 85.6 \\
    LingBot-VLA~\citep{wu2026pragmatic} & 88.6 & 86.7 & 87.7 \\
    \midrule
    \rowcolor{tablepurple} \textbf{\our-2B} & 87.9 & 87.2 & 87.6 \\
    \rowcolor{tablepurple} \textbf{\our-4B} & \textbf{88.7} & \textbf{88.4} & \textbf{88.6} \\
    \bottomrule
  \end{tabular}
\end{table}

\begin{table}
  \caption{\textbf{Results on VLABench.} SR, PS, and IS denote success rate, progress score, and intention score. The Average columns report arithmetic means across the five tracks when all five values are available. Dashes denote unreported measurements.}
  \label{tab:exp_vlabench}
  \centering
  \small
  \setlength{\tabcolsep}{8pt}
  \begin{tabular}{l ccc ccc ccc}
    \toprule
    \multicolumn{1}{c}{} & \multicolumn{3}{c}{\textbf{In-Distribution}}
     & \multicolumn{3}{c}{\textbf{Cross-Category}}
     & \multicolumn{3}{c}{\textbf{Common Sense}} \\
    \cmidrule(lr){2-4} \cmidrule(lr){5-7} \cmidrule(lr){8-10}
    \textbf{Method} & {\small SR} & {\small PS} & {\small IS}
     & {\small SR} & {\small PS} & {\small IS}
     & {\small SR} & {\small PS} & {\small IS} \\
    \midrule
    $\pi_0$~\citep{black2024pi_0}
     & 47.0 & 62.7 & 67.8 & 21.2 & 33.6 & 44.0 & 29.1 & 43.0 & 54.9 \\
    ACoT-VLA~\citep{zhong2026acot}
     & - & 66.1 & 79.8 & - & 38.9 & 54.1 & - & 37.8 & 52.3 \\
    $\pi_0$-FAST~\citep{pertsch2025fast}
     & 56.2 & 66.8 & 72.4 & 31.0 & 38.2 & 47.8 & 38.0 & 48.6 & 56.8 \\
    X-VLA~\citep{zheng2026x}
     & - & 67.8 & - & - & 25.1 & - & - & 48.2 & - \\
    $\pi_{0.5}$~\citep{intelligence2025pi05}
     & 65.4 & 77.8 & 80.4 & 38.2 & 49.7 & 52.0 & 43.9 & \textbf{57.3} & \textbf{60.0} \\
     \midrule
    \rowcolor{tablepurple} \textbf{\our-2B} & 58.5 & 71.2 & 81.2 & 30.3 & 40.1 & 53.3 & 26.2 & 39.9 & 59.8 \\
    \rowcolor{tablepurple} \textbf{\our-4B} & \textbf{67.8} & \textbf{80.3} & \textbf{83.9} & \textbf{48.0} & \textbf{60.7} & \textbf{68.2} & \textbf{44.6} & 53.8 & 54.1 \\
    \bottomrule
    \noalign{\vspace{6pt}}
    \toprule
    \multicolumn{1}{c}{} & \multicolumn{3}{c}{\textbf{Semantic Instruction}}
     & \multicolumn{3}{c}{\textbf{Unseen Texture}}
     & \multicolumn{3}{c}{\textbf{Average}} \\
    \cmidrule(lr){2-4} \cmidrule(lr){5-7} \cmidrule(lr){8-10}
    \textbf{Method} & {\small SR} & {\small PS} & {\small IS}
     & {\small SR} & {\small PS} & {\small IS}
     & {\small SR} & {\small PS} & {\small IS} \\
    \midrule
    $\pi_0$~\citep{black2024pi_0}
     & 17.3 & 38.7 & 58.0 & 32.2 & 42.5 & 50.6 & 29.4 & 44.1 & 55.0 \\
    ACoT-VLA~\citep{zhong2026acot}
     & - & 39.6 & 56.8 & - & 54.6 & 74.6 & - & 47.4 & 63.5 \\
    $\pi_0$-FAST~\citep{pertsch2025fast}
     & 35.0 & 45.0 & 59.4 & 39.0 & 49.0 & 56.8 & 39.8 & 49.5 & 58.6 \\
    X-VLA~\citep{zheng2026x}
     & - & 63.1 & - & - & - & - & - & 51.1 & - \\
    $\pi_{0.5}$~\citep{intelligence2025pi05}
     & 48.2 & 64.2 & 67.0 & 44.9 & \textbf{62.3} & 65.0 & 48.1 & 62.3 & 64.9 \\
     \midrule
    \rowcolor{tablepurple} \textbf{\our-2B} & 34.0 & 51.0 & \textbf{73.0} & 37.1 & 50.7 & \textbf{74.5} & 37.2 & 50.6 & 68.4 \\
    \rowcolor{tablepurple} \textbf{\our-4B} & \textbf{53.0} & \textbf{69.8} & 71.1 & \textbf{45.9} & 62.1 & 72.9 & \textbf{51.9} & \textbf{65.3} & \textbf{70.0} \\
    \bottomrule
  \end{tabular}
\end{table}

\subsection{Real-World Robot Tasks}
We evaluate post-trained policies on six tasks spanning bimanual coordination, tool use, and mobile manipulation. The tasks are distributed across two dual-arm platforms: \textbf{Marvin-WUJI}, a stationary dual-arm robot equipped with dexterous hands, and \textbf{Astribot-S1}, a wheeled mobile dual-arm robot equipped with parallel grippers. The demonstrations are collected through the teleoperation setup described in Appendix~\ref{app:teleop}, with per-task episode counts reported in Tab.~\ref{tab:teleop_data_stats}. The six tasks are defined as follows:

\textbf{Tasks performed on Marvin-WUJI (Dual-Arm Dexterous Hands):}
\begin{itemize}
    \item \textbf{(1) Plate-Placement:} The robot reaches toward a shelf, uses its dexterous hands to securely grasp a plate, and places it on the tabletop to the left of the oven, requiring precise spatial alignment and stable grasping of thin, fragile objects.
    \item \textbf{(2) Steak-Cooking:} The robot uses a spatula to flip a steak in a pan, requiring coordinated tool manipulation.
    \item \textbf{(3) Plate-Wiping:} A highly dexterous bimanual coordination task in which one hand grasps and stabilizes a plate while the other hand holds a cloth to wipe the plate's surface before returning the plate to the shelf, requiring coordinated manipulation.
\end{itemize}

\textbf{Tasks performed on Astribot-S1 (Wheeled Mobile Dual-Arm Grippers):}
\begin{itemize}
    \item \textbf{(4) Table-Cleaning:} The mobile robot collects scattered items on the table into a box and then places the box on a shelf, combining object collection with mobile manipulation.
    \item \textbf{(5) Steak-Serving:} The robot approaches the dining area, uses its dual grippers to stably lift a plate containing a cooked steak, and accurately serves it onto the designated spot on the dining table, integrating mobile navigation with precise placement.
    \item \textbf{(6) Plate-Retrieval:} Following the meal, the robot identifies and grasps used, dirty plates left on the dining table, collects the associated cutlery, and returns the items to their designated positions.
\end{itemize}

\begin{table}[t]
    \centering
    \caption{\textbf{Success rates on real-world robotic manipulation tasks}. Values are closed-loop success rates in percent on six tasks after robot-specific post-training. Each setting is evaluated with 35 trials per task. Bold indicates the highest observed rate per task among the listed comparisons.}
    \label{tab:real_world_eval}
    \small
    \setlength{\tabcolsep}{7pt}
    \begin{tabular}{lcccccc}
        \toprule
        \multirow{2}{*}{Model} & \multicolumn{3}{c}{\textbf{Marvin-WUJI}} & \multicolumn{3}{c}{\textbf{Astribot-S1}} \\
        \cmidrule(lr){2-4} \cmidrule(lr){5-7}
         & \shortstack{Plate \\ Placement} & \shortstack{Steak \\ Cooking} & \shortstack{Plate \\ Wiping} & \shortstack{Table \\ Cleaning} & \shortstack{Steak \\ Serving} & \shortstack{Plate \\ Retrieval} \\
        \midrule
        $\pi_0$~\citep{black2024pi_0} & 80.00 & 34.29 & 71.43 & 22.86 & 37.14 & 45.71 \\
        $\pi_{0.5}$~\citep{intelligence2025pi05} & 82.86 & 28.57 & 77.14 & 25.71 & 40.00 & 45.71 \\
        GR00T-N1.7~\citep{bjorck2025gr00t} & 74.29 & 28.57 & 74.29 & 17.14 & 45.71 & 37.14 \\
        \midrule
        \rowcolor{tablepurple} \textbf{\our-2B} & 80.00 & 34.29 & 68.57 & 34.29 & 42.86 & 48.57 \\
        \rowcolor{tablepurple} \textbf{\our-4B} & \textbf{85.71} & \textbf{42.86} & \textbf{82.86} & \textbf{40.00} & \textbf{57.14} & \textbf{54.29} \\
        \bottomrule
    \end{tabular}
\end{table}

\subsection{Results}

\paragraph{Latent-action representation.}
Tab.~\ref{tab:exp_lary} reports low-level control regression on LARY~\citep{nie2026lary}, with extended analyses under the relative protocol reported in Appendix~\ref{app:extended_ablations_lam}. In the absolute setting, the un-assembled encoder output (\lam (Raw Features)) achieves average MSE $0.14$, but its RoboCOIN error ($0.24$) remains above DINOv3 ($0.22$). The full descriptor achieves the lowest average MSE among the listed embodied LAMs in both target settings: $0.27$ absolute and $0.38$ relative. Relative MSE is lower than DreamDojo ($0.55$), Moto ($0.77$), and LAPA ($0.93$) on average, with gains on VLABench, RoboCOIN, and AgiBotWorld-Beta. DreamDojo remains better on relative CALVIN ($0.35$ versus $0.40$), and Moto has the lowest absolute CALVIN error in that block. Thus the representation improves average control predictability without uniformly dominating every dataset. Relative targets emphasize displacement prediction, while absolute targets also reward encoding the currently visible pose.

\paragraph{LIBERO and robustness.}
Tab.~\ref{tab:exp_libero} reports average LIBERO success rates of $98.3\%$ and $98.8\%$ for the 2B and 4B variants. The 4B result is the highest reported average in the table, exceeding Xiaomi-Robotics-0 by $0.1$ percentage point; this small difference should be interpreted as a reported score difference rather than evidence of statistical significance. The 4B variant leads Spatial and ties the highest Goal score, while other methods lead Object and Long. The 2B variant also exceeds 4B on Long ($97.8\%$ versus $97.0\%$), so increasing model size does not improve every suite.

On LIBERO-Plus (Tab.~\ref{tab:exp_liberoplus}), the same LIBERO-finetuned policies obtain $82.9\%$ and $84.7\%$ without further benchmark-specific adaptation. The 4B variant has the highest total among the listed comparisons, narrowly above $\pi_{0.5}$ ($84.4\%$), and leads the language, lighting, noise, and layout perturbations. However, $\pi_{0.5}$ performs better on camera and robot-state perturbations, while StarVLA leads background perturbations. These results show strong overall robustness with remaining weaknesses under viewpoint and initial-state changes.

\paragraph{Bimanual manipulation.}
RoboTwin 2.0 Native results (Tab.~\ref{tab:exp_robotwin}) demonstrate strong performance, with the 4B variant achieving the highest success rates among all listed methods. Specifically, \our-4B achieves $88.7\%$ on Clean and $88.4\%$ on Randomized scenes, yielding an average success rate of $88.6\%$. The 2B variant also performs competitively, scoring $87.9\%$ Clean, $87.2\%$ Randomized, and $87.6\%$ on average. The 4B variant exceeds the 2B variant by $0.8$ percentage points in Clean and $1.2$ in Randomized, yielding a $1.1$-point gain in average success rate. Notably, the 4B variant's Clean score surpasses LingBot-VLA ($88.6\%$), and its Randomized score exceeds both LingBot-VLA ($86.7\%$) and StarVLA-FAST ($83.2\%$), establishing a new leading result on this benchmark under the Native protocol. Because the Native protocol includes randomized demonstrations during training, these results assess performance within that protocol rather than zero-shot transfer to randomized scenes.

\begin{table}
  \caption{\textbf{Policy variants on LIBERO and LIBERO-Plus.} Values are success rates in percent. Avg. is the arithmetic mean of the two benchmark scores. All rows, including the latent-action references, are trained on the same approximately 1,000-hour subset and are therefore directly comparable; they differ from the full-corpus results in Tabs.~\ref{tab:exp_libero} and~\ref{tab:exp_liberoplus}.}
  \label{tab:ablation}
  \centering
  \small
  \begin{tabular}{l c c c}
    \toprule
    \textbf{Method} & \textbf{LIBERO} & \textbf{LIBERO-Plus} & \textbf{Avg.}  \\
    \midrule
    \our-2B-QwenVL~\citep{bai2025qwen3} & 98.1 & 78.8 & 88.5 \\
    \our-4B-QwenVL~\citep{bai2025qwen3} & 98.6 & 79.7 & 89.2 \\
    \our-2B w/o camera slots & 98.5 & 81.9 & 90.2 \\
    \our-4B w/o camera slots & 98.3 & 82.9 & 90.6 \\
    \our-2B Shared vector representation & 98.7 & 80.5 & 89.6 \\
    \our-4B Shared vector representation & \textbf{99.2} & 83.5 & 91.4 \\
    \our-2B Domain-specific projection & 99.0 & 82.1 & 90.6 \\
    \our-4B Domain-specific projection & 99.1 & 83.6 & 91.4 \\   
    \our-2B-LAPA~\citep{ye2025latent} & 98.2 & 80.2 & 89.2 \\   
    \our-4B-LAPA~\citep{ye2025latent} & 98.0 & 81.3  & 89.7 \\   
    \midrule
    \rowcolor{tablepurple} \textbf{\our-2B (RynnLAM)} & 98.1 & 82.3 & 90.2 \\
    \rowcolor{tablepurple} \textbf{\our-4B (RynnLAM)} & 98.7 & \textbf{84.2} & \textbf{91.5} \\
    \bottomrule
  \end{tabular}
\end{table}

\paragraph{Semantic grounding.}
Tab.~\ref{tab:exp_vlabench} reports five VLABench tracks. The 4B variant achieves the highest listed success rate on each track and arithmetic-mean SR, PS, and IS of $51.9$, $65.3$, and $70.0$, respectively. Its cross-category SR is $48.0\%$, compared with $38.2\%$ for $\pi_{0.5}$. The gains are not uniform across auxiliary scores: $\pi_{0.5}$ leads common-sense PS and IS, and narrowly leads unseen-texture PS. The 2B variant has lower average SR and PS than $\pi_{0.5}$ but higher average IS, illustrating that interpreting an instruction and completing the corresponding manipulation can remain separate challenges.

\begin{figure}[t]
  \centering
    \includegraphics[width=\linewidth]{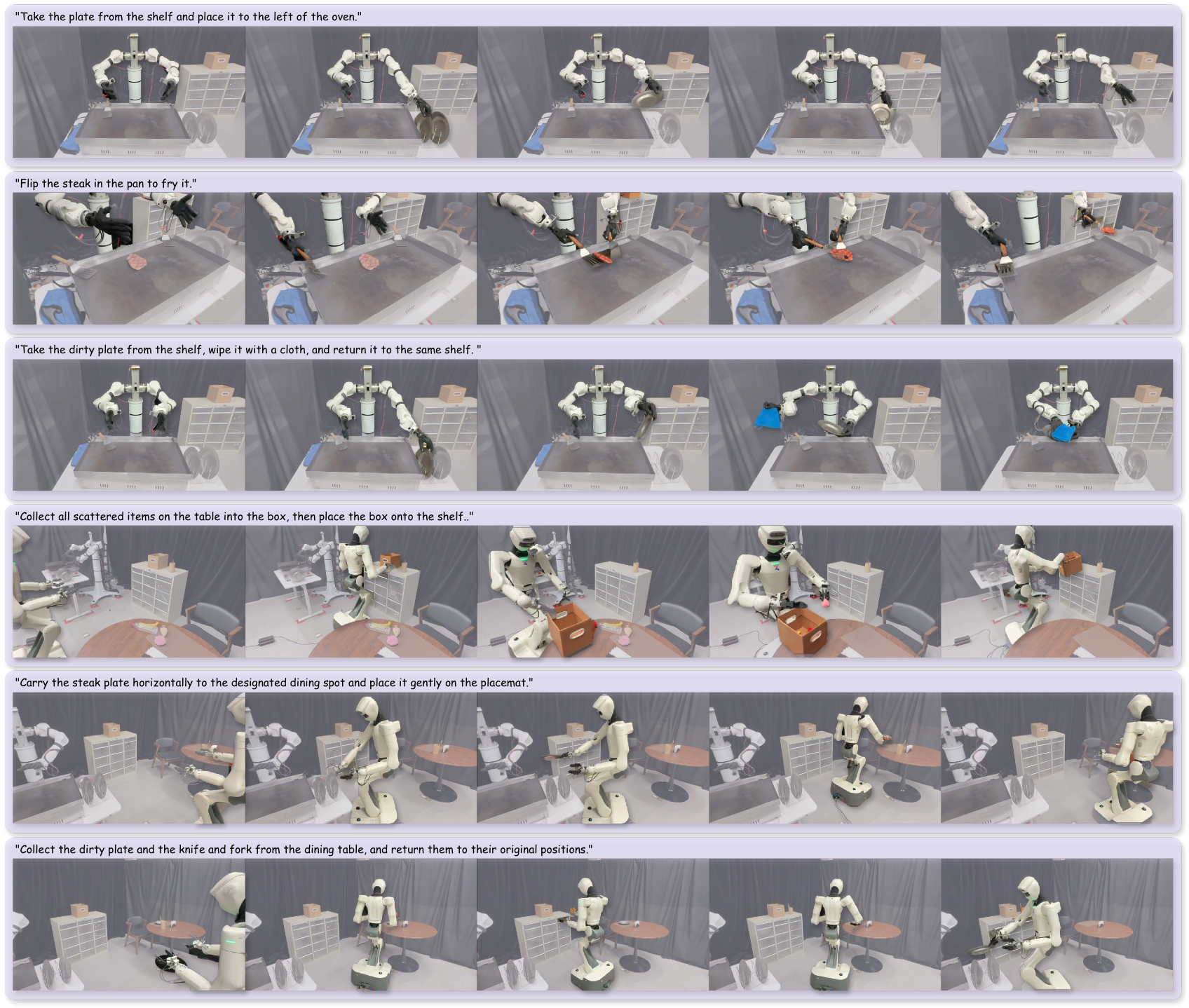}
  \caption{\textbf{Post-training evaluation on real-world tasks.} Each row shows the language instruction and representative execution frames for one task on the corresponding platform.}
  \label{fig:vis-real}
\end{figure}

\paragraph{Real-world manipulation.}
Tab.~\ref{tab:real_world_eval} reports closed-loop results on the two platforms. The 4B variant has the highest observed success rate among the listed methods on all six tasks. Relative to $\pi_{0.5}$, Table-Cleaning improves from $25.71\%$ to $40.00\%$ and Steak-Serving from $40.00\%$ to $57.14\%$. The 2B variant provides a more mixed comparison: it improves Table-Cleaning and Plate-Retrieval over the listed baselines, but trails $\pi_{0.5}$ on Plate-Placement and Plate-Wiping and trails GR00T-N1.7 on Steak-Serving. These results support the utility of the pretrained policy after target-robot post-training, rather than zero-shot deployment on an unseen embodiment. Fig.~\ref{fig:vis-real} shows representative executions.

\begin{figure}[htbp]
  \centering
    \includegraphics[width=\linewidth]{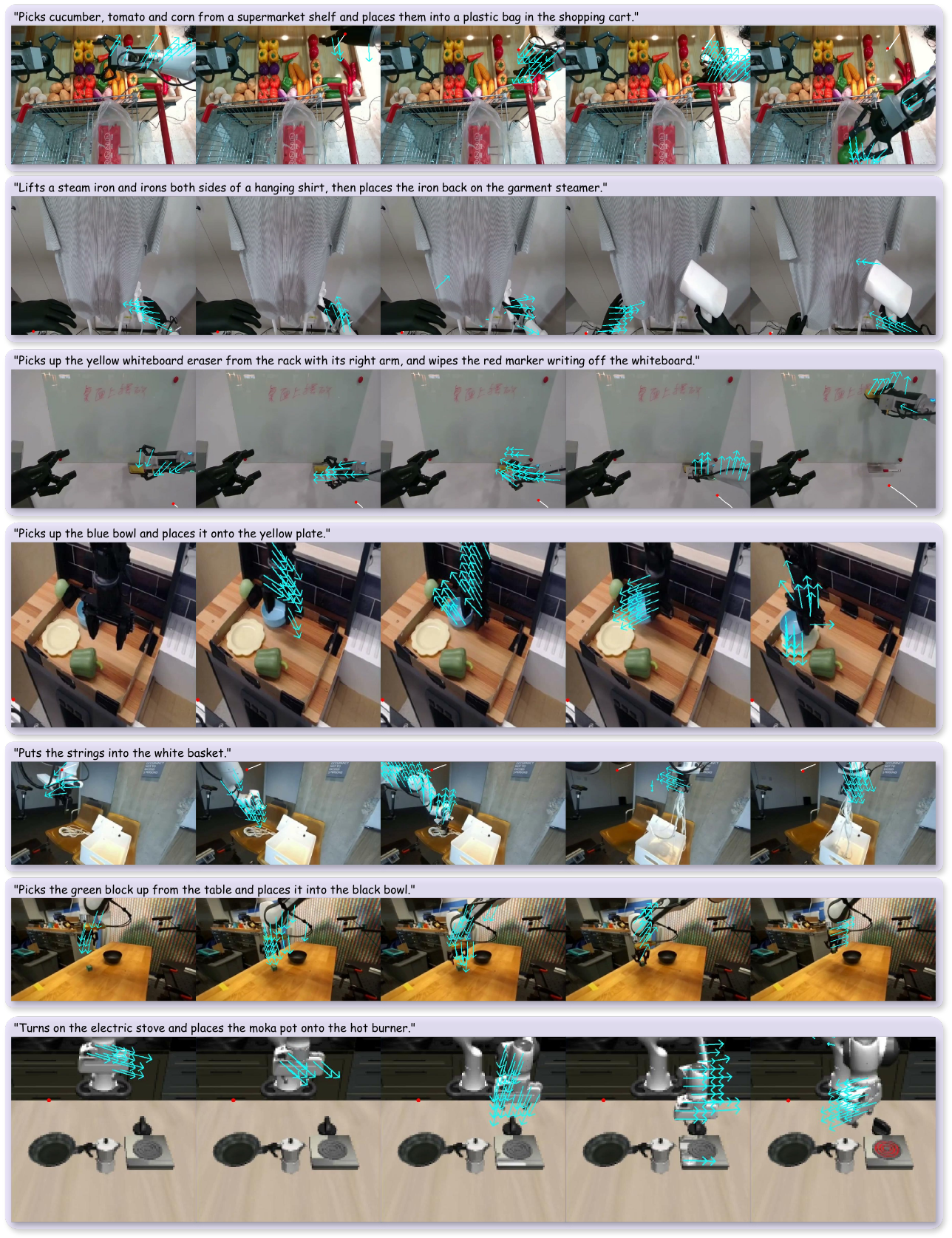}
  \caption{\textbf{Qualitative visualization of 3D scene flow} decoded from latent actions predicted by \our. Given the current observation and language instruction, \our predicts a latent-action chunk, which is fed into the \lam flow decoder to visualize the implied 3D motion.}
  \label{fig:vis-flow}
\end{figure}

\paragraph{Qualitative interpretation of the latent space.}
Fig.~\ref{fig:vis-latent-similarity} illustrates related motion patterns among retrieved latent neighbors across domains. To visualize policy-predicted latents, we pass a predicted chunk to the frozen \lam 3D-flow decoder (Fig.~\ref{fig:vis-flow}). The decoded fields provide a qualitative interpretation of the motion implied by the latent predictions. They do not independently establish metric physical accuracy or executable-action consistency.

\subsection{Component Studies and Data Scaling}

\subsubsection{\lam}
We further ablate the main designs of \lam in Tab.~\ref{tab:exp_lary}. Since the absolute setting can be partially affected by the currently visible end-effector pose, we mainly discuss the relative setting, while the absolute results show consistent trends.

\noindent
\textbf{End-to-end training.}
Training with a frozen encoder increases the relative average MSE from $0.38$ to $0.52$, and increases the absolute average MSE from $0.27$ to $0.37$. This gap supports updating the backbone during latent-action learning to improve the representation's utility for the evaluated control-regression tasks.

\noindent
\textbf{3D flow grounding.}
Removing 3D flow supervision increases the relative average MSE from $0.38$ to $0.50$, and increases the absolute average MSE to $0.37$. The error increases on all four datasets in both settings. This supports geometric scene-flow supervision as a useful signal for learning control-predictive descriptors; the probe does not directly measure the metric accuracy of recovered 3D motion.

\noindent
\textbf{Camera-motion decoupling.}
Removing the camera branch or the GRL objective increases the relative average MSE to $0.47$ and $0.49$, respectively, compared with $0.38$ for the full model. The corresponding absolute averages are both $0.32$, compared with $0.27$ for the full model. The error increases in both target settings support camera-pose supervision and the adversarial objective as useful components for control predictability. These ablations do not independently establish viewpoint invariance, which is assessed separately through downstream perturbation results.

\noindent
\textbf{Motion-token residual forcing.}
Removing motion-token residual forcing causes the largest degradation in the relative setting, increasing the average MSE from $0.38$ to $0.57$. Its absolute average also increases to $0.37$. The increase on every dataset supports residual future-feature prediction as a useful objective for preserving control-relevant information in the compact motion tokens. It is consistent with the intended role of the bottleneck, although the probe does not identify the specific shortcut used by a variant without this objective.

\begin{figure}[t]
  \centering
    \includegraphics[width=0.7\linewidth]{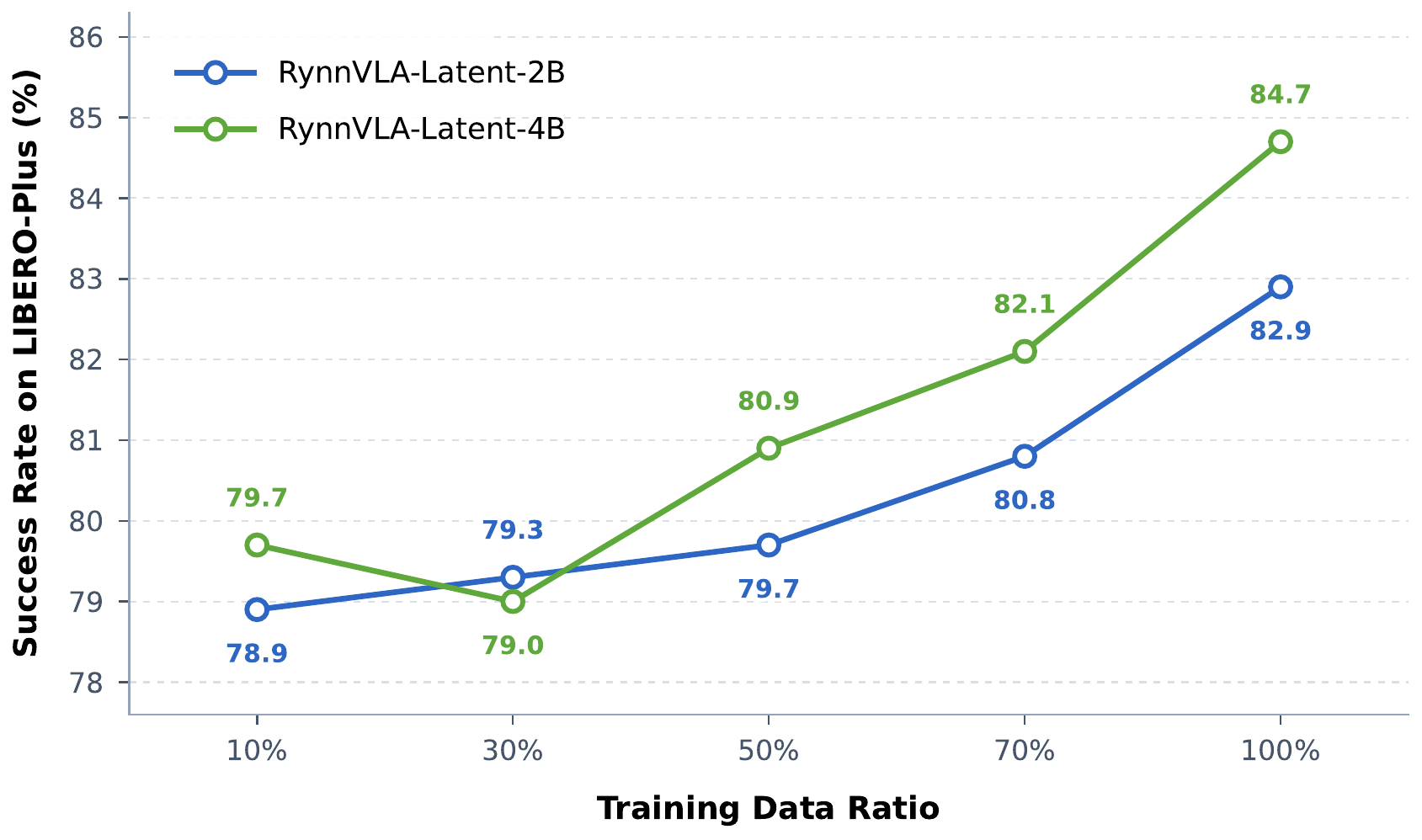}
  \caption{\textbf{Scaling of latent-action pretraining.} We pretrain \textbf{\textcolor[HTML]{2E66C4}{\our-2B}} and \textbf{\textcolor[HTML]{5FA83C}{\our-4B}} with 10\%, 30\%, 50\%, 70\% and 100\% of the latent-action pretraining corpus, and evaluate the resulting policies on LIBERO-Plus~\citep{fei25libero-plus}. Both models achieve their highest success rates with the full corpus, although improvements are not strictly monotonic across all data fractions.}
  \label{fig:scaling}
\end{figure}

\subsubsection{\our Policy Variants}

To isolate the effects of model architecture and action-supervision design, we conduct controlled comparisons on an approximately 1,000-hour subset drawn from RoboCOIN~\citep{wu2025robocoin}, EgoDex~\citep{hoque2025egodex}, EPIC-KITCHENS~\citep{damen2020epic}, and Something-Something~\citep{goyal2017something}. All variants in Tab.~\ref{tab:ablation}, including the RynnLAM-based reference models, use this subset rather than the full 30.1K-hour corpus. We therefore distinguish these controlled configuration comparisons from the full-scale results reported in Tabs.~\ref{tab:exp_libero} and \ref{tab:exp_liberoplus}.

\noindent
\textbf{Latent Actions as Cross-Embodiment Action Alignment.}
We compare RynnLAM-based supervision against two strategies for aligning heterogeneous action spaces: a shared canonical action vector with masking~\citep{yuan2026qwen,intelligence2025pi05} and embodiment-specific input and output projections~\citep{agarwal2026cosmos3,zheng2026x}. For human videos without native action labels, both explicit alignment baselines use motion-capture-derived wrist poses and fingertip positions as pseudo-action supervision, with implementation details provided in Appendix~\ref{app:action_alignment_baselines}. As shown in Tab.~\ref{tab:ablation}, under the same approximately 1,000-hour pretraining subset, RynnLAM-based pretraining achieves higher LIBERO-Plus success than both alternatives: 82.3\% versus 80.5\%/82.1\% for 2B, and 84.2\% versus 83.5\%/83.6\% for 4B. In contrast, the explicit action-alignment variants achieve higher LIBERO success. These results suggest that latent-action supervision can favor robustness to distribution shifts, while explicit native-action alignment remains competitive for in-distribution task execution.

\noindent
\textbf{Backbone, camera slots, and latent-action targets.}
Under the shared training subset, the RynnBrain-based models outperform their Qwen-VL counterparts on LIBERO-Plus by 3.5 and 4.5 percentage points for 2B and 4B, respectively. Removing canonical camera slots reduces LIBERO-Plus success by 0.4 and 1.3 percentage points, while replacing RynnLAM descriptors with LAPA targets reduces it by 2.1 and 2.9 percentage points. These results support the combined design of the vision-language backbone, multi-view conditioning interface, and continuous latent-action supervision, with the most consistent differences appearing under perturbation-based evaluation.

\noindent
\textbf{Scaling Behavior.}
We investigate the impact of latent-action pretraining data scale by training RynnVLA-Latent with 10\%, 30\%, 50\%, 70\%, and 100\% of the video corpus and evaluating the resulting policies on LIBERO-Plus under the same downstream protocol. As shown in Fig.~\ref{fig:scaling}, both 2B and 4B models achieve their highest success rates with the full pretraining corpus, demonstrating an overall benefit from scaling up heterogeneous video supervision. While the 2B model improves monotonically across the evaluated data fractions, the 4B model exhibits a slight decrease from 10\% to 30\% before improving with larger data scales. These results suggest that increasing latent-action pretraining data generally improves downstream robustness within the evaluated range, although the relationship is not strictly monotonic for every model size.

\section{Conclusion}
We presented a framework for latent-action pretraining of vision-language-action policies from heterogeneous interaction video. \lam combines residual feature prediction, camera-compensated 3D-flow pseudo-labels, contrastive warping, and camera-pose decoupling to produce continuous transition descriptors without native robot-action labels. \our learns to predict these descriptors through flow matching, then reuses its pretrained backbone and action expert while retargeting the output space to robot commands.

The evaluation supports this interface at two levels. Frozen probes show that the descriptor is predictive of low-level controls, with average relative-action MSE $0.38$ on LARY. After native-action post-training, the 4B policy achieves $98.8\%$ on LIBERO, $84.7\%$ on LIBERO-Plus, $51.9\%$ average success on the five reported VLABench tracks, and $88.6\%$ under the RoboTwin 2.0 Native protocol. It also attains the best success rate among the compared baselines on all six real-world tasks across stationary dexterous and mobile gripper platforms. Performance remains task-dependent, and RoboTwin scores remain below the strongest listed baselines. Together with the data-scaling trend, these findings support shared transition supervision as a practical route to policy pretraining across incompatible control spaces.

\paragraph{Limitations.}
Depth, camera poses, and flow are estimated pseudo-labels, so geometric supervision does not guarantee calibrated metric motion or complete separation of camera and manipulation information. Low-level control probes measure predictability, and decoded flow provides qualitative interpretation, rather than proof of causal action identifiability. Policy performance varies across tasks and perturbations, and deployment still requires embodiment-specific action-labeled demonstrations. The current results also do not establish a general scaling law or isolate pretraining-data volume from all differences in optimization exposure. Future work will improve geometric consistency, viewpoint robustness, and transfer with less target-robot supervision.

\newpage
\bibliographystyle{assets/plainnat}
\bibliography{paper}

\newpage
\beginappendix
\section{Benchmark Details}
\label{app:benchmark_details}

\subsection{Latent Action Representation Quality}
We first evaluate the latent action representation itself. Rather than judging latent
actions by downstream manipulation success, which entangles representation quality with
the rest of the VLA stack, the LARY benchmark~\citep{nie2026lary} probes the frozen
representation directly. We adopt its \emph{low-level control regression} track, which
decodes end-effector trajectories from a frozen representation through a lightweight
MLP action expert, scored by MSE across CALVIN~\citep{mees2022calvin} and
VLABench~\citep{zhang2025vlabench} (7-DoF Franka), RoboCOIN~\citep{wu2025robocoin}
(12-DoF bimanual) and AgiBotWorld-Beta~\citep{bu2025agibot} (16-DoF AgiBot G1); we leave
aside its semantic classification track, which scores category separability rather than
physical grounding. The official targets are \emph{absolute} end-effector states in a
fixed base frame. At the short stride ($s{=}5$) used to form each image pair, the currently
visible pose can help predict these targets. Absolute-target performance may
therefore reflect static pose information as well as transition information. We therefore also report a
\emph{relative} setting that re-expresses the same targets as displacements from the
end-effector state at the first frame, leaving the probe, splits, stride and metric
untouched, so that the probe emphasizes displacement prediction rather than absolute pose reconstruction. We compare
against three of the paradigms of~\citep{nie2026lary}: general semantic encoders
(V-JEPA~2~\citep{assran2025vjepa2}, DINOv3~\citep{simeoni2025dinov3}), generative pixel
encoders (Wan2.2~\citep{wan2025}, FLUX.2-dev~\citep{flux-2-2025}), and Embodied LAMs
(LAPA~\citep{ye2025latent}, UniVLA~\citep{bu2025univla},
villa-X~\citep{chen2026villa}), plus Moto~\citep{chen2025moto} and
DreamDojo~\citep{gao2026dreamdojo} evaluated under the identical protocol.

\noindent 
\textbf{LIBERO and LIBERO-Plus.}
LIBERO~\citep{liu2023libero} is a language-conditioned tabletop benchmark on a Franka Emika
Panda. Its four suites---\emph{Spatial}, \emph{Object}, \emph{Goal}, \emph{Long}, 10
tasks each with 50 human-teleoperated demonstrations---isolate spatial grounding, target
disambiguation, goal following, and long-horizon chaining. We fine-tune a single policy
jointly across all 40 tasks and evaluate each suite over 500 rollouts (50 per task).
LIBERO-Plus~\citep{fei25libero-plus} reuses the same 40 tasks as a robustness diagnostic,
perturbing seven factors---layout, viewpoint,
robot initial state, language, lighting, background, and sensor noise---over 10{,}030
curated episodes. We evaluate the LIBERO-finetuned checkpoint without additional LIBERO-Plus adaptation to measure robustness under these perturbations.

\noindent 
\textbf{VLABench.}
VLABench~\citep{zhang2025vlabench} returns to a fixed Franka Panda but shifts the
difficulty from control to grounding: task categories spanning primitive and
composite interactions, built in MuJoCo over 2{,}164 objects in 163 categories, with instructions
written to convey intent implicitly rather than by template. We fine-tune on
the benchmark's official primitive-task split and evaluate on the fixed
episode sets of the benchmark's public tracks, which hold the tasks constant while
varying what the policy must resolve to locate the target: \emph{in-distribution},
\emph{cross-category}, \emph{common-sense}, \emph{semantic-instruction}, and \emph{unseen-texture}. We omit the benchmark's cross-task
track, whose train/test partition is left to the user and whose numbers are therefore not
comparable across papers.

\noindent 
\textbf{RoboTwin 2.0.}
RoboTwin 2.0~\citep{chen2025robotwin} contains 50 bimanual manipulation tasks, including handovers, cooperative lifting, and stabilization followed by manipulation. The benchmark provides 731 object instances in 147 categories and supports multiple dual-arm embodiments. We use ALOHA-AgileX and the Native protocol, with 50 clean and 500 randomized demonstrations per task (27,500 in total). A single policy is trained jointly across all 50 tasks and evaluated in clean and randomized scenes. Randomization covers clutter, lighting, background, tabletop height, and language. Because randomized demonstrations are included during training, the two evaluation domains measure performance within this training protocol rather than zero-shot transfer from clean training to randomized scenes. Results for this protocol are reported in Tab.~\ref{tab:exp_robotwin}.

\section{Teleoperation Setup}
\label{app:teleop}
\begin{figure}
  \centering
    \includegraphics[width=0.9\linewidth]{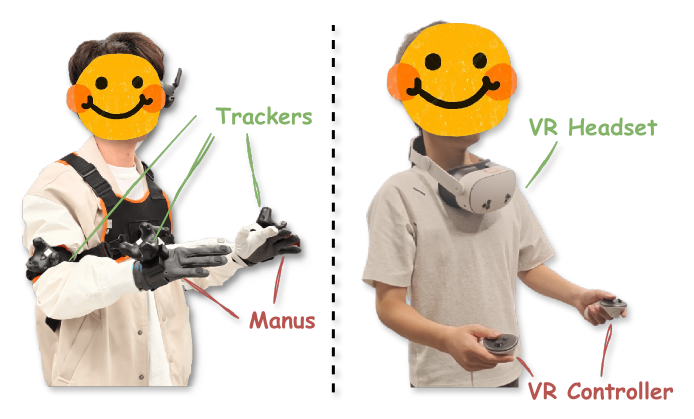}\\
  \caption{\textbf{Marvin-WUJI demonstration collection.} Motion trackers control the dual robot arms, and data gloves control the dexterous hands.}
  \label{fig:operator}
\end{figure}

For Marvin-WUJI, we collect demonstrations through the teleoperation system shown in Fig.~\ref{fig:operator}. The hardware setup consists of dual Tianji 7-DOF robotic arms and dual Wuji 20-DOF dexterous hands, yielding a total of 54 degrees of freedom.                            

For arm control, the human operator wears five HTC Vive trackers (mounted on the chest, both wrists, and both upper arms). The system computes wrist-to-chest relative transforms at 100-120 Hz and feeds them into a Pinocchio-based inverse kinematics solver running in a separate process. The resulting joint commands are further smoothed by a Ruckig trajectory generator with velocity, acceleration, and jerk constraints before being sent to the robot arms at 200 Hz.                  

For hand control, the operator wears Manus data gloves. The raw glove signals are converted to a 21-point MediaPipe hand skeleton format and retargeted to the 20-DOF Wuji hand joint space via a dedicated retargeting module, with an exponential moving average filter applied for motion smoothing.

\section{Details of Action-Alignment Baselines}
\label{app:action_alignment_baselines}

To compare our latent-action-based cross-embodiment alignment with conventional action-alignment strategies, we implement two baselines commonly used in VLA pretraining.

\subsection{Shared Vector Representation}

Following prior VLA research~\citep{yuan2026qwen,intelligence2025pi05}, we map heterogeneous state and action spaces from different data sources and embodiments into a unified $85$-dimensional canonical vector representation. The layout is designed for dual-arm embodiments. For single-arm robots, only the first half of each bilateral slot group is populated. The vector is organized into the following semantic slots:

\begin{enumerate}[leftmargin=*, nosep]
    \item \textbf{Joint Position (0--13):} dual-arm joint positions, with $7$ dimensions allocated to each arm.

    \item \textbf{EEF Pose (14--31):} Cartesian end-effector position ($3$ dimensions) and orientation represented using a continuous $6$-D rotation representation ($6$ dimensions) for each arm.

    \item \textbf{Gripper Position (32--33):} one gripper-position dimension for each arm.

    \item \textbf{Dexterous Hand (34--73):} $20$ dimensions reserved for each dexterous hand.

    \item \textbf{Torso (74--77):} $4$ dimensions for torso degrees of freedom.

    \item \textbf{Head (78--80):} $3$ dimensions for head and neck motion.

    \item \textbf{Base (81--84):} $4$ dimensions reserved for mobile-base degrees of freedom.
\end{enumerate}

Different embodiments populate different subsets of this unified vector according to their available degrees of freedom. Unused dimensions are excluded from the training loss through a per-dimension binary mask, preventing spurious supervision on structurally absent action dimensions.

\subsection{Domain-specific Projection}

Following prior work~\citep{agarwal2026cosmos3,zheng2026x}, this variant retains a shared action expert but replaces the common input and output projections with a bank of domain-specific projections. Each physical robot embodiment is assigned a domain identifier, which selects the corresponding projection parameters for that sample. In this design, the main policy remains shared across embodiments, while the lightweight boundary projections capture embodiment-specific differences in action parameterization and distribution.

\subsection{Human Egocentric Datasets}

For human egocentric datasets, we treat human motion as a separate embodiment type when using domain-specific projections. These videos carry no native control labels; wrist poses and the positions of five fingertips are mapped to the corresponding slots of the unified vector.

\subsection{Delta Actions and Normalization}

During training, all action slots use a delta-action representation: each action in an action chunk is expressed relative to the robot's current state. For normalization, the $6$-D end-effector rotation slots are left unchanged. All other action slots are normalized using the $q_{0.01}$ and $q_{0.99}$ quantiles computed from the training data.

\section{\our Dataset}

\noindent 
\textbf{Video captioning.}
We use Qwen3-VL~\citep{bai2025qwen3} to generate captions for the video data. Specifically, we leverage the model's strong video-language understanding capabilities to produce detailed, structured descriptions of each video clip. The videos are first sampled at a frame rate of 1 FPS and split into segments of 5 seconds. For each segment, we provide the following prompt to the model:

\begin{mdframed}[backgroundcolor=tablepurple!40, linewidth=0.9pt, linecolor=tablepurple, roundcorner=10pt]
\texttt{Please describe this video in detail. Include the following aspects:} \\ \smallskip
\texttt{1. The main subject and action in the video.} \\ \smallskip
\texttt{2. The environment and background.} \\ \smallskip
\texttt{3. Any objects and their interactions.} \\ \smallskip
\texttt{4. The overall scene context and atmosphere.} \\ \smallskip
\texttt{Provide a concise but comprehensive caption in one paragraph.}
\end{mdframed}                                                                                

The generation is performed with a maximum output length of 512 tokens and a temperature of 0.7 to balance creativity and coherence. The generated captions are then collected and stored in JSON format for downstream tasks.

\noindent 
\textbf{Optical flow annotation.}
We employ DPFlow~\citep{morimitsu2025dpflow}, a state-of-the-art optical flow estimation model. For each video, frame pairs are processed sequentially at native resolution, and the estimated flow fields are visualized via color encoding and saved as MP4 videos at 25 FPS.

\noindent 
\textbf{Depth annotation.}
We employ Depth Anything 3~\citep{lin2025depth}, specifically the \texttt{DA3NESTED-GIANT-\allowbreak LARGE-1.1} checkpoint, which provides dense per-frame depth predictions along with camera pose estimation. Each video is sampled at 30 FPS and processed at a working resolution of 392 pixels (short side, upper-bound resize).

\section{Extended Ablations of \lam}
\label{app:extended_ablations_lam}

To complement the main ablations, we further evaluate \lam under the relative low-level control protocol of the LARY benchmark~\citep{nie2026lary}. We focus on the relative action setting because it reduces reliance on reconstructing the currently visible absolute end-effector pose, and thus more directly reflects whether the frozen representation encodes motion rather than static scene appearance. We report the relative setting for these extended variants; absolute-action results are not included in this table. Unless otherwise stated, all retrained variants use the same data, budget, and probe protocol as the main table. The latent-composition rows are probe-only: they reuse the same released checkpoint and only change the slice of the latent passed to the probe.

\begin{table}[h]
  \caption{
    \textbf{Extended ablations on the LARY benchmark~\citep{nie2026lary}} (Relative Action, MSE, lower is better).
    All retrained variants use the same data, budget, and probe protocol as the main table. Avg. is the arithmetic mean of the displayed per-dataset MSE values, rounded to two decimal places.
  }
  \label{tab:exp_lary_extended}
  \centering
  \small
  \begin{tabular}{l ccccc}
    \toprule
    \textbf{Method} & \textbf{Avg.} & \textbf{CALVIN} & \textbf{VLABench} & \textbf{RoboCOIN} & \textbf{AgiBotWorld-B} \\
    \midrule

    \multicolumn{6}{l}{\emph{Training \& Objective Variants}} \\
    \lam w/ 2D flow & 0.56 & 0.44 & 0.44 & 0.69 & 0.67 \\
    \lam w/o contrastive feat. & 0.57 & 0.44 & 0.45 & 0.70 & 0.68 \\
    \lam w/o feat. prediction & 0.63 & 0.49 & 0.50 & 0.78 & 0.74 \\
    \lam w/ $z^{m}$ from latent tokens & 0.79 & 0.64 & 0.66 & 0.94 & 0.92 \\
    \midrule
    \rowcolor{tablepurple}
    \textbf{\lam} & \textbf{0.38} & \textbf{0.40} & \textbf{0.05} & \textbf{0.42} & \textbf{0.64} \\ 

    \midrule
    \multicolumn{6}{l}{\emph{Latent Composition (probe-only)}} \\
    $z^{a}$ only (64d) & 0.93 & 0.82 & 0.96 & 0.97 & 0.95 \\
    $z^{c}$ only (32d) & 0.95 & 0.88 & 0.97 & 0.98 & 0.96 \\
    $z^{m}$ only (512d) & 0.54 & 0.42 & 0.42 & 0.66 & 0.65 \\
    \bottomrule
  \end{tabular}
\end{table}

Tab.~\ref{tab:exp_lary_extended} reports the results. The first block studies training and objective variants. \textbf{\lam w/ 2D flow} replaces the 3D scene-flow decoder with a 2D appearance-flow decoder; together with the \emph{w/o 3D flow} variant in the main table, it compares no-flow, 2D-flow, and 3D-flow supervision. The full 3D-flow model has lower average error than either variant; adding 2D-flow supervision does not improve over the no-flow variant, so an additional motion-prediction objective alone does not explain the gain. \textbf{\lam w/o contrastive feat.} removes the anti-shortcut contrastive feature objective and falls back to the plain motion-weighted cosine loss, testing the utility of contrasting warped features against unchanged source features. \textbf{\lam w/o feat. prediction} disables the future feature-prediction branch entirely, measuring how much of the representation quality is driven by next-frame feature prediction versus the latent-specific objectives. \textbf{\lam w/ $z^{m}$ from latent tokens} compresses $z^{m}$ from the latent tokens instead of the raw target-frame features, removing the direct feature-space pathway into the bottleneck; it degrades the average MSE substantially, supporting the direct target-feature pathway for preserving information useful to the control-regression probe.

The second block probes $z^{a}$, $z^{c}$, and $z^{m}$ separately using the same checkpoint. Motion tokens yield the lowest average MSE among the individual components ($0.54$), but remain worse than the full descriptor ($0.38$). Thus $z^{m}$ carries substantial control-relevant information, while the combined descriptor gives the best average result. The global action and camera latents alone achieve $0.93$ and $0.95$, respectively, and are less predictive under this probe. These slice comparisons establish differences in probe utility; they do not by themselves prove a complete semantic disentanglement of the representation.

Together, these studies support the complete objective and the direct target-feature pathway for control regression. The component probes additionally show that the full descriptor is more predictive than any individual slice. These findings explain the choice of descriptor used for policy pretraining, while downstream evaluations separately assess its utility after adaptation to native robot controls.

\section{Real-World Data Collection Details}

\begin{table}[t]
\centering
\caption{Statistics of the real-world dataset used for \our training.}
\label{tab:teleop_data_stats}
\begin{tabular}{lp{10cm}c}
\toprule
\textbf{Task Name} & \textbf{Description} & \textbf{Episodes} \\ \midrule
Plate-Placement & Take the plate from the shelf and place it to the left of the oven. & 500 \\
 Steak-Cooking & Flip the steak in the pan to fry it. & 800 \\
Plate-Wiping & Take the dirty plate from the shelf, wipe it with a cloth, and return it to the same shelf. & 500 \\
 Table-Cleaning & Collect all scattered items on the table into the box, then place the box onto the shelf. & 850 \\
Steak-Serving & Carry the steak plate horizontally to the designated dining spot and place it gently on the placemat. & 600 \\
Plate-Retrieval & Collect the dirty plate and the knife and fork from the dining table, and return them to their original positions.  & 600 \\
\midrule
\rowcolor{headerpurple!60} \textbf{Total} & & \textbf{3,850} \\ \bottomrule
\end{tabular}
\end{table}

Operator motion tracking provides the control interface for collecting the demonstrations. For policy post-training, recorded robot observations and native actions are aligned in time, and action chunks are expressed in the target robot's control space. Operator skeletons are not the latent-action targets used during policy pretraining: those targets are extracted from video by \lam.

In total, we collected 3,850 high-quality demonstration episodes. The corpus covers all six evaluated tasks; these counts describe the collected demonstrations across the two platforms. The specific tasks and their respective data distributions are detailed in Tab.~\ref{tab:teleop_data_stats}.

\end{document}